\pdfoutput=1
\documentclass[11pt]{article}

\usepackage[final]{acl}

\usepackage{times}
\usepackage{latexsym}

\usepackage{booktabs}
\usepackage{graphicx}

\usepackage{amsmath}
\usepackage{amssymb}

\usepackage{fvextra}
\usepackage{varwidth}

\usepackage{float}

\usepackage[table]{xcolor}

\usepackage{subcaption}
\usepackage{listings}
\usepackage{inconsolata}  % 一个等宽字体，适合展示代码和prompt

\usepackage{verbatim}
\usepackage{fancyvrb}
\usepackage{etoolbox}

\lstdefinestyle{promptstyle}{
    basicstyle=\ttfamily\small,
    breaklines=true,
    breakatwhitespace=false,
    breakindent=0pt,
    columns=fullflexible,
    keepspaces=true,
    showstringspaces=false,
    mathescape=true,
    escapeinside={(*@}{@*)},
    linewidth=\linewidth,
    breakindent=0pt,
    literate={[}{[}{1}
             {]}{]}{1}
             {\_}{\_}{1}
             {\{}{\{}{1}
             {\}}{\}}{1},
    keywordstyle=\color{blue},
    stringstyle=\color{red},
    commentstyle=\color{green!70!black}
}

\DefineVerbatimEnvironment{PromptVerbatim}{Verbatim}{
    fontsize=\small,
    baselinestretch=0.85,
    commandchars=\\\{\}  % 允许在verbatim中使用部分LaTeX命令
}

\usepackage[T1]{fontenc}
\usepackage[utf8]{inputenc}

\usepackage{microtype}

\usepackage{inconsolata}

\usepackage{graphicx}
\usepackage{multirow}
\usepackage{makecell}
\title{\textbf{\textsc{AskToAct}}: Enhancing LLMs Tool Use via \\ Self-Correcting Clarification}

\author{
  \textbf{Xuan Zhang\textsuperscript{1}}, 
  \textbf{Yongliang Shen\textsuperscript{1}\textsuperscript{†}}, 
  \textbf{Zhe Zheng\textsuperscript{1}}, 
  \textbf{Linjuan Wu\textsuperscript{1}}, 
  \textbf{Wenqi Zhang\textsuperscript{1}}, \\
  \textbf{Yuchen Yan\textsuperscript{1}}, 
  \textbf{Qiuying Peng\textsuperscript{2}}, 
  \textbf{Jun Wang\textsuperscript{2}}, 
  \textbf{Weiming Lu\textsuperscript{1}\textsuperscript{†}} \\
  \textsuperscript{1}Zhejiang University \\
  \textsuperscript{2}OPPO Research Institute \\
  \{zhangx0513, syl, luwm\}@zju.edu.cn, 
  \{pengqiuying, wangjun7\}@oppo.com
}

\begin{document}

\maketitle

\begingroup
\renewcommand\thefootnote{\textsuperscript{†}}
\footnotetext{Corresponding author.}
\endgroup

\begin{abstract}

% Large language models (LLMs) have demonstrated remarkable capabilities in tool learning. In real-world scenarios, user queries are often ambiguous and incomplete, requiring effective clarification. However, existing interactive clarification approaches face two critical limitations: reliance on manually constructed datasets and lack of error correction mechanisms during multi-turn clarification. 
Large language models (LLMs) have demonstrated remarkable capabilities in tool learning. In real-world scenarios, user queries are often ambiguous and incomplete, requiring effective clarification. However, existing interactive clarification approaches face two critical limitations: reliance on manually constructed datasets, which inherently constrains training data scale and diversity, and lack of error correction mechanisms during multi-turn clarification, leading to error accumulation that compromises both accuracy and efficiency.
% We present \textbf{\textsc{AskToAct}}, which addresses these challenges by exploiting the structural mapping between queries and their tool invocation solutions. Our key insight is that tool parameters naturally represent explicit user intents. By systematically removing key parameters from queries while retaining them as ground truth, we enable automated construction of high-quality intent clarification data. We further enhance model robustness by fine-tuning on error-correction augmented data using selective masking mechanism, enabling dynamic error detection and recovery during clarification interactions. 
We present \textbf{\textsc{AskToAct}}, which addresses these challenges by exploiting the structural mapping between queries and their tool invocation solutions. Our key insight is that tool parameters naturally represent explicit user intents. By systematically removing key parameters from queries while retaining them as ground truth, we enable automated construction of high-quality training data. We further enhance model robustness through error-correction pairs and selective masking, enabling dynamic error detection during clarification interactions.
Comprehensive experiments demonstrate that \textbf{\textsc{AskToAct}} significantly outperforms existing approaches,  achieving above 57\% accuracy in recovering critical unspecified intents and enhancing clarification efficiency by an average of 10.46\% while maintaining high accuracy in tool invocation.  Our framework exhibits robust performance across different model architectures and successfully generalizes to entirely unseen APIs without additional training, achieving performance comparable to GPT-4o with substantially fewer computational resources.

\end{abstract}

\begin{figure*}[ht]
    \centering
    \begin{minipage}[t]{0.33\textwidth}
        \centering
        \includegraphics[trim=0.3cm 10.9cm 17.4cm 0.8cm, clip, width=\textwidth]{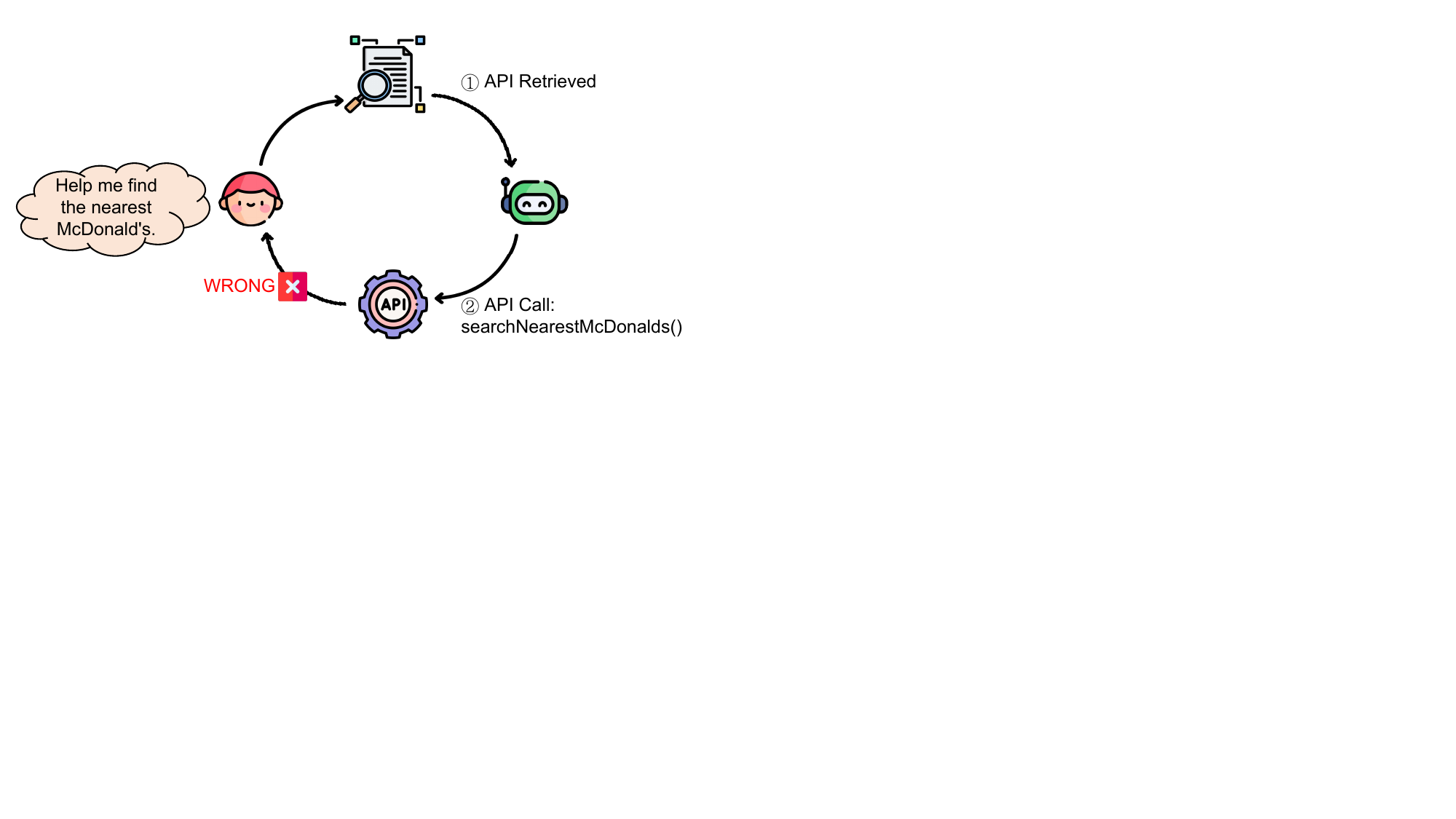}
        {\footnotesize{(a)}}
    \end{minipage}%
    % \hfill
    \begin{minipage}[t]{0.33\textwidth}
        \centering
        \includegraphics[trim=0.3cm 10.9cm 17.4cm 0.8cm, clip, width=\textwidth]{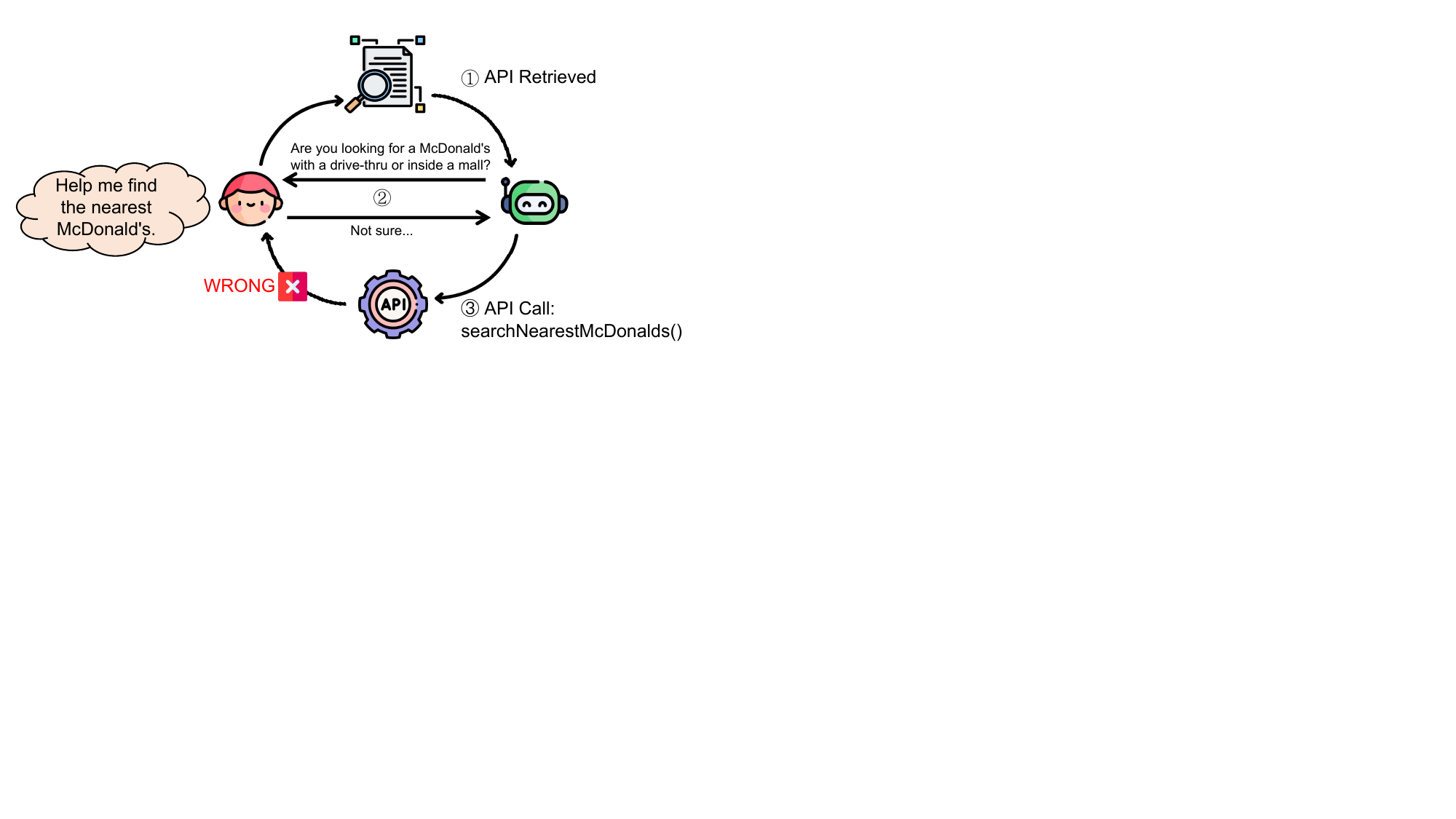}
         {\footnotesize{(b)}}
    \end{minipage}%
    % \hfill
    \begin{minipage}[t]{0.33\textwidth}
        \centering
        \includegraphics[trim=0.3cm 12.8cm 14.7cm 2.9cm, clip, width=\textwidth]{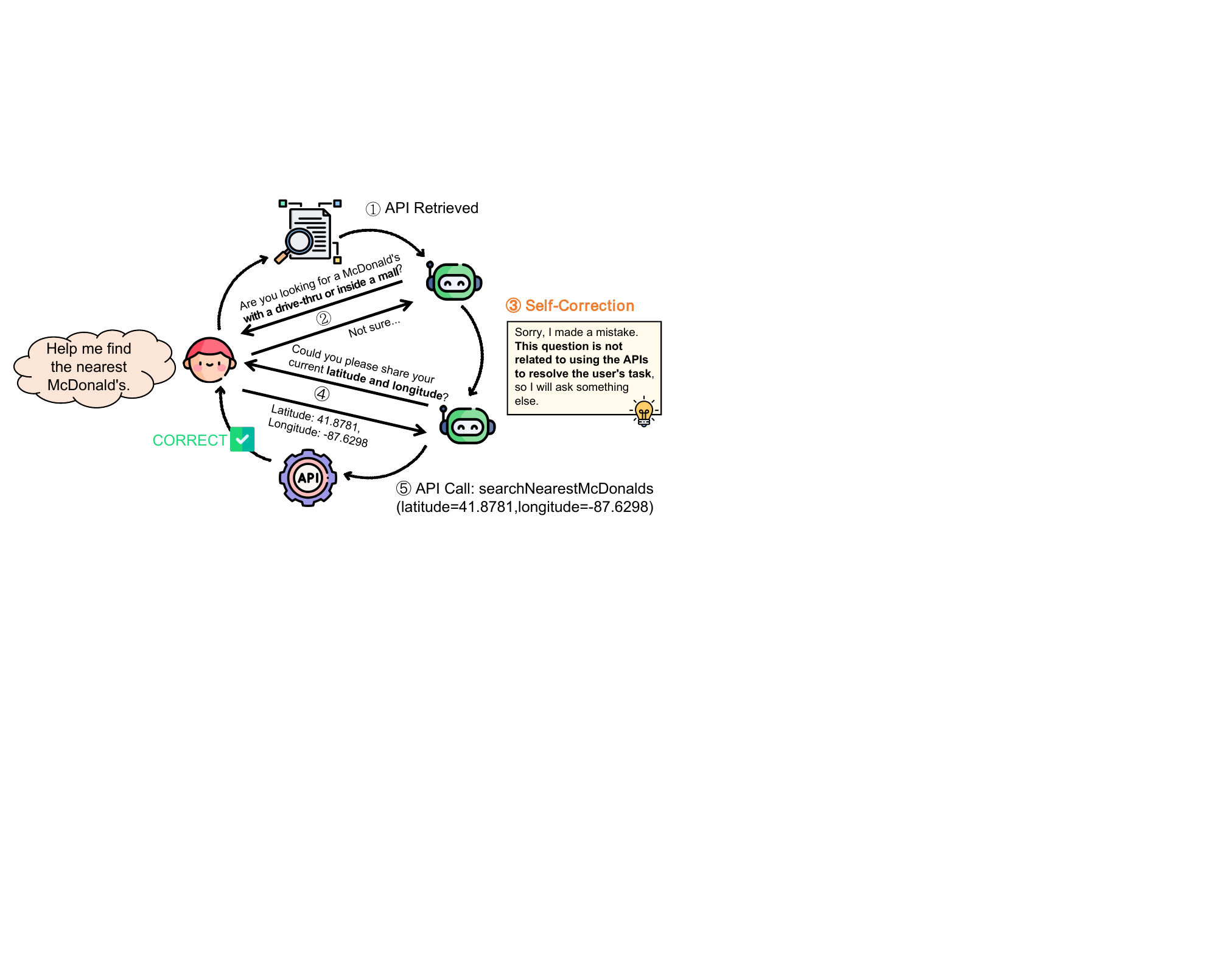}
        {\footnotesize{(c)}}
    \end{minipage}
    \caption{Comparison of query handling approaches: (a) direct API calls without clarification, (b) basic clarification without error recovery, and (c) our self-correcting \textbf{\textsc{AskToAct}} framework.}
    \label{fig:clarification-capability-compare}
\end{figure*}

\section{Introduction}

Large language models (LLMs) have demonstrated remarkable capabilities in various tasks, from code generation to complex reasoning 
 \cite{nakano2021webgpt,chen2021evaluating,komeili-etal-2022-internet, NEURIPS2022_9d560961}. A particularly promising direction is their ability to interact with external tools through API calls, which significantly expands their practical applications 
 \cite{NEURIPS2023_d842425e,NEURIPS2023_8fd1a81c,qin2023toolllm, shim2025tooldial}. This has inspired numerous frameworks focusing on tool-augmented LLMs, including Toolformer \cite{NEURIPS2023_d842425e}, ToolLLaMA \cite{qin2023toolllm}, and Gorilla \cite{patil2023gorilla}.

However, current tool learning frameworks \cite{li-etal-2023-api,song2023restgptconnectinglargelanguage, NEURIPS2023_d842425e, qin2023toolllm} operate under an idealistic assumption that user queries are always explicit and unambiguous. This diverges significantly from real-world scenarios where users often provide incomplete, ambiguous, or imprecise queries. Such unspecified queries pose unique challenges in tool learning scenarios, as API calls require precise parameters and cannot tolerate ambiguity \cite{wang2024learning}. When faced with unspecified queries, LLMs tend to either arbitrarily generate missing parameters or remain unknown, leading to potential risks in tool invocation.

This raises a critical research question: \textit{How can we enhance LLMs' ability to handle unspecified queries in tool learning scenarios while ensuring accurate and reliable tool invocation?} Recent works  \cite{zhang2023clarify, qian-etal-2024-tell, wang2024learning} have introduced interactive clarification approaches, but face two fundamental limitations.
First, they rely heavily on manually constructed datasets for training \cite{qian-etal-2024-tell, wang2024learning}. Creating these datasets requires human annotators to craft queries and clarifications, a process that inherently limits scale and diversity. The resulting datasets capture only a narrow range of ambiguity patterns, reducing their effectiveness with diverse real-world queries. 
Second, as shown in Figure~\ref{fig:clarification-capability-compare}, these approaches lack robust error handling during multi-turn clarification. Existing models train on datasets with only perfect clarification sequences. In reality, models often request already-provided information, follow irrelevant paths, or miss unspecified details. Without error recovery training, these issues accumulate throughout dialogues, reducing efficiency and compromising tool invocation quality.

We propose \textbf{\textsc{AskToAct}}, a self-correcting clarification framework that systematically addresses these limitations. Our key insight is that tool parameters naturally represent explicit user intents, creating an opportunity for automated data generation. We develop an automated pipeline that strategically removes key parameters from complete queries in existing datasets, generating diverse unspecified queries with built-in ground truth. Using these queries, we construct rich clarification dialogues demonstrating effective intent elicitation. To enable robust error handling during interactions, we augment training data with carefully designed error-correction pairs that simulate realistic mistakes and their solutions. We implement selective masking during training to prevent learning negative patterns while enhancing error detection abilities.

% \endgroup

% \begingroup
% \setlength{\parskip}{2pt} 

Through comprehensive experiments, we demonstrate that \textbf{\textsc{AskToAct}} achieves several significant improvements: (1) correctly identifies unspecified queries and recovers more than 57\% of critical unspecified intents, while significantly enhances clarification efficiency by an average of 10.46\% compared to the base model; (2) achieves strong performance in end-to-end tool invocation, with over 81\% tool selection accuracy and over 68\% parameter resolution accuracy;
(3) exhibits robust performance across different model architectures, and successfully generalizes to entirely unseen APIs;
and (4) delivers performance comparable to GPT-4o while requiring substantially fewer computational resources.
% \endgroup

% \begingroup
% \setlength{\parskip}{2pt} 

Our work makes three main contributions:

% \endgroup

\begin{itemize}
    \item We introduce an automated pipeline for generating high-quality intent clarification datasets, addressing the scalability limitations of manual annotation.
    \item We develop a self-correction mechanism that enables models to dynamically detect and correct potential errors during clarification interactions. 
    \item Our experimental results demonstrate that our method not only achieves state-of-the-art (SOTA) performance but also shows strong generalization ability when handling queries requiring the use of unseen APIs.
\end{itemize}

\begin{figure*}[t]
    \centering
    \includegraphics[trim=2.7cm 3.2cm 3.4cm 3cm, width=0.98\linewidth]{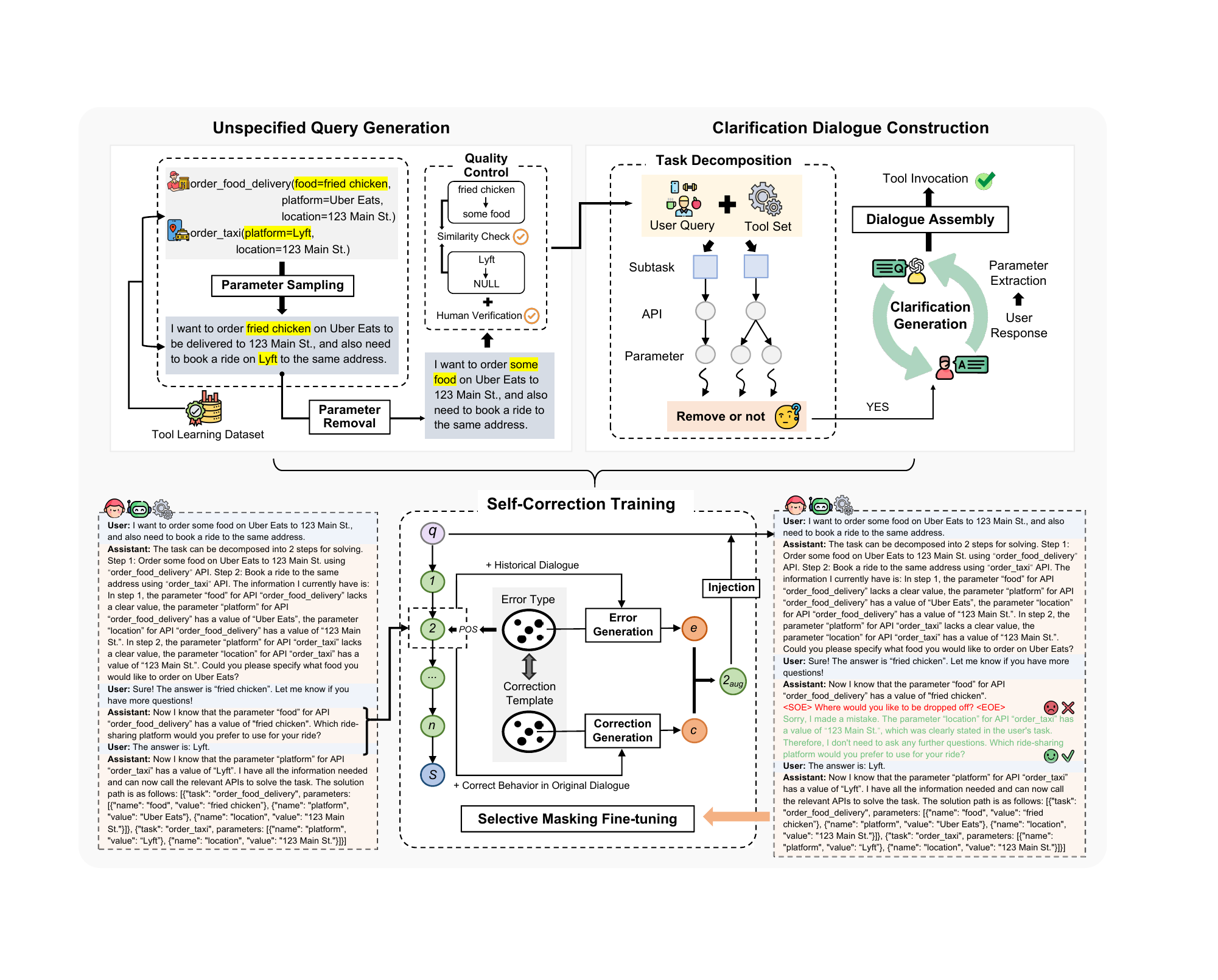}
\caption{Overview of \textbf{\textsc{AskToAct}} framework. 
Top: Dataset construction pipeline, consisting of (1) unspecified query generation by selecting and removing key parameters (e.g., "fried chicken" and "Lyft") from original queries, and (2) dialogue construction through task decomposition, clarification generation, and dialogue assembly.
% Top: Dataset construction pipeline, showing (1) unspecified query generation through parameter selection and removal (\textit{e.g.}, "fried chicken" and "Lyft") from original queries and (2) dialogue construction with task decomposition, clarification generation and dialogue assembly. 
Bottom: Self-correction training through dialogue augmentation and selective masking fine-tuning.}
    % 结合图的具体例子解释()
    \label{fig:framework}
\end{figure*}

% \begingroup
% \vspace{-6pt}
% \section{Method}
% \vspace{-2pt}
% \endgroup

\section{Method}

Tool learning faces a fundamental challenge: while API calls require precise parameters, real-world queries are often ambiguous. 
To bridge this gap, we propose \textbf{\textsc{AskToAct}}, a self-correcting clarification framework. Our method consists of two key components: 
% We present \textbf{\textsc{AskToAct}}, a self-correcting clarification framework that addresses this challenge through two key components: 
(1) an automated data construction pipeline for generating diverse intent calrification data (§\ref{sec:dialogue_construction}), and (2) a self-correction training paradigm for dynamic error detection and correction (§\ref{sec:training_paradigm}). The core insight is that tool parameters naturally represent explicit user intents, making them ideal anchors for both data generation and error correction. Figure~\ref{fig:framework} illustrates the overall framework architecture.

\subsection{Intent Clarification Dataset Curation}
\label{sec:dialogue_construction}

The foundation of our method is a systematic pipeline for constructing multi-turn clarification  data. As shown in Figure~\ref{fig:framework}, the pipeline proceeds in two steps: generating unspecified queries and subsequently constructing clarification dialogues.

\subsubsection{Unspecified Query Generation}

A key challenge in building intent clarification systems is obtaining realistic examples of ambiguous queries paired with their complete intents. We address this through a novel reverse-engineering approach that leverages existing tool learning datasets. 
Each instance in these datasets consists of a fully specified query $q$ and a corresponding tool invocation solution $S = \{(f_i, P_i) \mid i = 1, \dots, n\}$, where $f_i$ denotes the API and $P_i = \{p_i^1, p_i^2, \dots\}$ represents its parameter set. We systematically transform $q$ into an unspecified query $q'$ while preserving the ground truth information necessary for subsequent dialogue construction and evaluation.
% Given a fully specified query $q$ and its corresponding tool invocation solution $S = \{(f_i, P_i) | i = 1,...,n\}$, where $f_i$ represents the API and $P_i = \{p_i^1, p_i^2, ...\}$ represents its parameter set, we systematically transform it into an unspecified query $q'$ while preserving the ground truth information necessary for subsequent dialogue construction and evaluation.

% \begingroup
% \setlength{\parskip}{0pt} 

% \vspace{-5pt}

\paragraph{Parameter Sampling} 
The first step in our pipeline is to determine which parameters in $S$ to remove from the original query $q$. We implement a stratified sampling approach that enables sampling across different API domains and parameter counts. For each query, we randomly select parameters according to one of four complexity levels: (1) fully specified, where all parameters are retained, (2) single-API single-parameter, where one parameter from one API call is removed, (3) single-API multi-parameter, where multiple parameters from the same API call are removed, and (4) multi-API multi-parameter, where parameters are removed across multiple API calls. This stratification ensures our dataset captures the full spectrum of query ambiguity encountered in real-world scenarios, from basic single-slot ambiguities to complex multi-faceted ambiguities.

% \endgroup

% \begingroup
% \setlength{\parskip}{2pt} 

\paragraph{Parameter Removal} 
Once the removed parameter set $P'$ is determined, we apply two complementary strategies to transform the original query $q$ into an unspecified form. The first strategy, \textit{complete removal}, entirely eliminates parameter values from $q$ while preserving grammatical integrity. The second strategy, \textit{semantic abstraction}, replaces specific parameter values with abstract expressions that necessitate further clarification.
For each parameter $p \in P'$, we maintain a mapping $M: p \rightarrow v'$, where $v'$ represents the transformed value after parameter removal or abstraction. 
By recording the values before and after the transformation, we can more precisely track how explicit user intent becomes unspecified  during the process of unspecified query generation.
This record plays an important role in quality control, helping us ensure the quality of the generated query. Also, it guides subsequent dialogue construction and provides ground truth for evaluation. 
The implementation details and the format of the transformation record are provided in Appendix~\ref{sec:unspecified_query_generation} and Appendix~\ref{sec:unspecified_json}, respectively.

% \endgroup

% Once the removed parameter set $P'$ is determined, we employ two complementary strategies to transform the original query while maintaining natural language fluency. The first strategy, \textit{complete removal}, eliminates parameter values while preserving grammatical structure. The second strategy, \textit{semantic abstraction}, replaces specific values with abstract expressions that necessitate further clarification. 
% For each parameter $p \in P'$, we maintain a mapping $M: p \rightarrow v'$ tracking the transformation in the query $q'$, where $v'$ represents the transformed value after parameter removal or abstraction. 
% % As shown in Figure~\ref{fig:framework}, if "fried chicken" is  changed to "some food" in the query, the mapping would track this transformation ($p = \text{"food"}$, $v' = \text{"fried chicken"}$). 
% By recording the values before and after the transformation, we can more intuitively observe the changes in user intent during the process of unspecified query generation. This record plays an important role in subsequent filtering and quality control, helping us ensure the validity and accuracy of the data. Also, it guides subsequent dialogue construction and provides ground truth for evaluation. We leverage GPT-4o to ensure contextually appropriate transformations, with detailed prompts provided in Appendix~\ref{sec:unspecified_query_generation}.

\paragraph{Quality Control} 

\renewcommand{\arraystretch}{1.05}
\begin{table*}[!h]
\centering
\small
\resizebox{0.85\textwidth}{!}{
% \begin{tabular}{lccc}
% \toprule
% & \textbf{Daily Life} & \textbf{Hugging Face} & \textbf{Multimedia} \\
% \midrule
% Fully-Specified & 275  & 16  & 5 \\
% Sing-API-Sing-Param & 761  & 140  & 107 \\
% Sing-API-Multi-Param & 488  & -  & - \\
% Multi-API-Multi-Param & 1857  & 20  & 15 \\
% \midrule
% Total & 3381 & 176 & 127 \\
% \bottomrule
% \end{tabular}
% }

    % \centering \small
\begin{tabular}{lcccccc}
    \toprule
    \textbf{Dataset} & \textbf{Train}  & \textbf{Test} & \textbf{Avg. No. APIs} & \textbf{Avg. No. Params} & \textbf{Avg. No. Unspecified Intents} \\
    \midrule
    xlam-IC & 29,821 &  4,456 & 1.58 & 2.49 & 1.32\\
    Taskbench-IC & - & 984 & 1.75 & 2.35 & 1.30 \\
    \midrule
    \textbf{Total} & 29,821 & 5,440 & 1.59 & 2.49 & 1.32\\
    \bottomrule
\end{tabular}
}

%     \caption{Datasets Overview}
%     \label{tab:datasets}
% \end{table}

\caption{Datasets statistics. {xlam-IC} is generated from the xlam-function-calling-60k dataset~\cite{NEURIPS2024_61cce86d} which is used for training and testing.
{Taskbench-IC} is generated from Taskbench~\cite{shen2024taskbench} and is used exclusively for OOD testing.
Please refer to the Appendix~\ref{sec:dataset} for more details about both datasets.}

\label{tab:dataset_statistics}
\end{table*}

% To ensure generation quality, we employ a multi-stage verification process. We first compute semantic similarity between original values and their transformations using Sentence Transformer \cite{reimers-2019-sentence-bert}. Queries with similarity scores exceeding 0.85 are filtered out to ensure sufficient semantic alteration. We then verify structural validity to maintain grammatical correctness and natural flow. 
% During the human evaluation, we randomly selected 50 samples from our TASKBENCH and invited three domain experts to assess the quality of these samples. To ensure a fair and unbiased evaluation, all samples will be anonymized. We provide canonical samples for these experts to calibrate their criteria during the annotations, and calculate an average score of all experts' ratings as the final results. All results can be found in Table ~\ref{tab:quality_control}.
% We randomly sampled approximately 10\% of the generated unspecified queries (368 queries) for manual validation by three graduate students with NLP backgrounds. Each query was evaluated using the following criteria:
To ensure generation quality, we employ a dual-stage verification mechanism. We first compute semantic similarity between original values and their transformations using Sentence Transformer \cite{reimers-2019-sentence-bert}. Queries with similarity scores exceeding 0.95 are filtered out to ensure sufficient semantic alteration. Then, we  conduct human verification on  generated queries (see Appendix~\ref{sec:human_verification_1}).
% Then, we perform human verification on 400 randomly sampled queries. Three graduate students with NLP backgrounds independently assessed each query based on six criteria: Naturalness (fluency and linguistic coherence), Consistency (uniformity of transformation), Necessity (need for clarification), Complexity (difficulty of clarification), Diversity (range of parameter types and domains), and Acceptance Rate (overall acceptability). Results are shown in Table~\ref{tab:quality_control1}.
Through this process, we constructed 35,261 high-quality unspecified queries, as shown in Table~\ref{tab:dataset_statistics}.

% 对应修改图

\subsubsection{Clarification Dialogue Construction}
Based on the generated unspecified queries, we propose an automated method to construct training dialogue data that simulate multi-turn clarification. The dialogue construction  process—from task decomposition, clarification generation to final dialogue assembly—is essential for generating coherent and effective clarification interactions.
% Once unspecified queries are generated, we develop a systematic approach to construct training dialogues that demonstrate effective intent clarification through multi-turn interactions. 
% Our dialogue construction process starts from task decomposition to final dialogue assembly, ensuring each step contributes to coherent and effective clarification interactions.
\paragraph{Task Decomposition}
The foundation of effective clarification lies in identifying what information needs to be clarified. Given an unspecified query $q'$ and its tool invocation solution $S$, we first decompose the query into a sequence of subtasks. Each subtask corresponds to an API call in $S$. For each API call, we perform parameter analysis to identify two categories: (1) parameters already specified in $q'$, and (2) parameters requiring clarification. This structured decomposition guides the subsequent clarification process, ensuring all necessary information is systematically obtained.

\paragraph{Clarification Generation}
Building on the task decomposition, we generate interaction turns for each parameter that requires clarification, following the API call order defined in $S$. We construct each clarification turn through a three-step process, with the goal of maximizing clarification effectiveness and preserving natural conversational flow.
First, we generate a clarification question $q_c$ targeting the unspecified parameter.
Next, we simulate the user’s reply using diverse response templates that vary in verbosity and conversational tone.
Finally, we generate a confirmation statement summarizing the clarified information, which serves as an explicit reference for subsequent turns.

% A turn consists of three essential components. First, we utilize GPT-4o to generate contextually appropriate questions $q_c$ that specifically target unspecified parameters, ensuring precision and clarity. Second, we simulate user responses by selecting from a diverse template set that varies in verbosity and conversational style. Finally, we generate an explicit confirmation statement for each response, creating a clear record of the clarified information that serves as reference for subsequent turns.

\paragraph{Dialogue Assembly}
The final step brings together all components into a coherent dialogue structure. We assemble the generated elements sequentially while maintaining natural conversation flow through consistent reference to previously clarified information and smooth transitions between parameter-related clarifications. Special attention is paid to parameter interdependencies, ensuring that information is requested in a logical order that reflects real-world dialogue patterns. The dialogue concludes with the complete tool invocation solution $S$, providing a clear connection between the clarification process and its ultimate goal. Detailed templates and prompting strategies that support this assembly process are provided in Appendix~\ref{sec:clarification_generation}.

% 引入外部表格文件
% \input{tables/information-missing-levels}

\subsection{Self-Correction  Training}
\label{sec:training_paradigm}

While constructing high-quality training data is essential, the dynamic nature of clarification interactions requires models to detect and correct potential errors in real-time. We develop a systematic training paradigm that combines error-correction augmentation with specialized training strategies to enhance model robustness and enable self-correction.

% \subsubsection{Error Type Analysis}
\label{sec:error_type}

\paragraph{Error Type Analysis}
Through comprehensive  analysis of clarification interactions generated by the model in response to unspecified queries, we identify five primary error types that impair the clarification process in complementary ways. 
\textit{Clearly Stated Intent Clarification} occurs when the model requests explicitly stated information, creating unnecessary interaction turns. \textit{Imprecise Clarification}  is characterized by questions that lack specificity, often resulting in ambiguous user responses.  
\textit{Irrelevant Clarification} emerges when the model poses questions that diverge from the core intent. \textit{Redundant Clarification} arises when the model requests information that has been previously clarified. 
\textit{Incomplete Clarification} represents failure to identify all parameters requiring clarification, leading to incomplete tool invocation solutions. Understanding these patterns guides our error-correction strategy.

% 引入外部表格文件

% \begin{table}[t]
% \centering \small
% \begin{tabular}{lc}
% \toprule
% \textbf{Error Type} & \textbf{Count} \\
% \midrule
% Clearly Stated Intent Clarification & 1,495 \\
% Imprecise Clarification & 1,355 \\
% Irrelevant Clarification & 1,371 \\
% Redundant Clarification & 1,858 \\
% Incomplete Clarification & 2,867 \\
% \midrule
% Total & 8,946 \\
% \bottomrule
% \end{tabular}
% \caption{Error type distribution in augmented dialogues.}
% \label{tab:error-types}
% \end{table}

\begin{table}[t]
\centering \small
\begin{tabular}{lc}
\toprule
\textbf{Error Type} & \textbf{Count} \\
\midrule
Clearly Stated Intent Clarification & 2,481 \\
Imprecise Clarification &  2,298\\
Irrelevant Clarification & 2,251 \\
Redundant Clarification & 3,126 \\
Incomplete Clarification &  5,000 \\
\midrule
Total & 15,156 \\
\bottomrule
\end{tabular}
\caption{Error type distribution in augmented dialogues.}
\label{tab:error-types}
\end{table}

\subsubsection{Error-Correction Augmentation}

% Building upon this error analysis, we introduce an automated method to augment dialogues with error-correction pairs. For each dialogue $d$ in our dataset $D = {(q_i, d_i)}$, where $q_i$ represents the unspecified query and $d_i$ represents the corresponding clarification dialogue, we randomly select an error type $\tau_k$ and determine an appropriate injection position $pos$. We generate errors using two strategies: For semantic errors (Clearly Stated Intent, Imprecise, and Irrelevant), we employ GPT-4o with specialized prompts: $e = f_{\text{gpt}}(d, \tau_k, pos)$. For structural errors (Redundant and Incomplete), we implement rule-based algorithms: $e = f_{\text{rule}}(d, \tau_k, pos)$. The detailed implementation of these generation strategies is provided in Appendix~\ref{sec:error_generate}.

Building upon this error analysis, we introduce an automated method to augment dialogues with error-correction pairs. Given a dialogue $d$, we randomly select an error type $\tau_k$ and determine an injection position $pos$. We then generate  the error instance using two strategies.  For semantic errors (Clearly Stated Intent, Imprecise, and Irrelevant), we employ GPT-4o with specialized prompts: $e = f_{\text{gpt}}(d, \tau_k, pos)$. For structural errors (Redundant and Incomplete), we implement rule-based algorithms: $e = f_{\text{rule}}(d, \tau_k, pos)$. Implementation details are provided in Appendix~\ref{sec:error_generate}.

After generating error $e$, we construct a correction $c$ using a template specific to the error type $\tau_k$. The resulting correction $c$  explicitly states the error type and identifies the correct behavior as the content at position $pos$ in the original dialogue $d$.
 % After generating error $e$, we create correction $c$ by applying type-specific templates that identify the error and specify that the correct behavior is the one at position $pos$ in the original dialogue $d$. 
 We then inject these error-correction pair $(e, c)$ into the original dialogue $d$ at the predetermined position $pos$: $d' = \text{inject}(d, e, c, pos)$. 

To ensure the validity  of our error-correction augmentation method, we conduct human verification on   augmented dialogues (see Appendix~\ref{sec:human_verification_2}).
 % To ensure the quality and utility of our error-correction augmentation method, we perform human evaluation on 200 randomly sampled augmented dialogues. Three graduate students with NLP backgrounds independently assessed each dialogue based on five criteria: Naturalness (fluency and linguistic coherence), Plausibility (realism of the injected error-correction pair),  Usefulness (utility of the error-correction pair for intent clarification),  Diversity (variation in error manifestation and expression styles), and Acceptance Rate (overall acceptability). Results are shown in Table~\ref{tab:quality_control2}.
Through this systematic process, we generated 15,156 augmented dialogues, as shown in Table~\ref{tab:error-types}. 
 % The resulting dataset $D_{aug}$ comprises these augmented dialogues along with the original unmodified dialogues from $D$.  
 % We present the dialogue examples in Appendix~\ref{sec:reflection_template}.

\subsubsection{Selective Masking Fine-tuning}

To effectively utilize the augmented dialogues for training while preventing the model from learning error patterns, we implement a selective masking mechanism during fine-tuning. We introduce special tokens \textit{<SOE>} and \textit{<EOE>} to demarcate error segments, and mask these segments during loss computation. This approach allows the model to learn error detection and correction patterns while avoiding the reinforcement of error behaviors. Through this training process, we equip the model with the ability to dynamically identify potential errors and apply appropriate corrections during clarification interactions.

\section{Experiment}
\subsection{Experimental Settings}

% \paragraph{Dataset} Our dataset consists of two sub-dataset: xlam-IC and Taskbench-IC. The xlam-IC dataset is divided into training and test sets at a 10:1 ratio, while the other one serves as an exclusive out-of-distribution (OOD) test set to rigorously evaluate the model's cross-domain generalization capabilities.  

\paragraph{Training Details}
% 在这里说30%
We construct our training data from the xlam-IC dialogue dataset, where 30\% of the samples are replaced with error-correction augmented dialogues.  We explore two adaptation strategies for the Qwen2.5-7B-Instruct model: LoRA~\cite{hu2021loralowrankadaptationlarge}  and full-parameter fine-tuning.
More details are provided in Appendix~\ref{sec:training_details}.

% We fine-tune LLaMA-2-7B-Chat using LLaMA-Factory \cite{zheng2024llamafactory}, with a maximum sequence length of 4096 tokens. The training uses a global batch size of 16, an initial learning rate of 1.41e-5 with 0.1 warm-up ratio, and a cosine learning rate scheduler. The model is trained for 2 epochs on 4 RTX A6000 GPUs. 

\paragraph{Baselines}  
For comprehensive comparison, we evaluate representative tool-augmented LLMs, including xLAM-7b-fc-r (trained on the xlam-function-calling-60k dataset but without intent clarification) \cite{NEURIPS2024_61cce86d}, gorilla-openfunctions-v2 \cite{patil2023gorilla}, and ToolLLaMA-2-7b-v2 \cite{qin2023toolllm}, as well as an intent clarification model, Mistral-Interact \cite{qian-etal-2024-tell}. In addition, we evaluate major LLM series, including Mistral-7B-Instruct-v0.3, LLaMA (3-8B/70B-Instruct), Qwen (2.5-7B/72B-Instruct), DeepSeek-V3, Claude (3.5-Haiku/Sonnet), and GPT (3.5, 4, 4o-mini, 4o). All models use a standardized evaluation prompt (see Appendix~\ref{sec:prompt_for_evalutaion}).

% We evaluate against specialized tool learning models including xLAM-7b-fc-r \cite{NEURIPS2024_61cce86d}, gorilla-openfunctions-v2 \cite{patil2023gorilla} and ToolLLaMA-2-7b-v2 \cite{qin2023toolllm}. Also, we compare with well-trained intent clarified model Mistral-Interact \cite{qian-etal-2024-tell}. For comprehensive comparison, we include major LLM series: LLaMA (3-8B/70B-Instruct), Mistral (7B-Instruct-v0.3), Qwen (2.5-7B/72B-Instruct), DeepSeek-V3, Claude (3.5-Haiku/Sonnet), and GPT (3.5, 4, 4o-mini, 4o). All models use a standardized evaluation prompt (Appendix~\ref{sec:prompt_for_evalutaion}).

\subsection{Evaluation Framework}

We develop an automated framework for systematic evaluation on handling unspecified queries. The framework employs an LLM to simulate user behavior. During interactions, the user-simulating LLM judges whether clarification questions are relevant to unspecified intents, and either provides the necessary information or indicates that it is unavailable. 
To better capture the complexity of real-world human-LLM interactions,  we configure the user-simulating LLM with six personality types, each exhibiting different response characteristics. Implementation details are provided in Appendix~\ref{sec:user_prompt}.

\renewcommand{\arraystretch}{1.05}
\begin{table*}[!h]
\centering
\small
\resizebox{\textwidth}{!}{
\begin{tabular}{l ccccccc}
\toprule
\multirow{2}{*}{\centering \textbf{LLM}} & \multicolumn{4}{c}{\textbf{Intent Clarification Quality}} & \multicolumn{3}{c}{\textbf{Tool Invocation Accuracy}}  \\
\cmidrule(lr){2-5} \cmidrule(lr){6-8} 
  & \textbf{ICR}$\uparrow$ & \textbf{CE}$\uparrow$ & \textbf{CPS}$\uparrow$ & \textbf{IR}$\downarrow$ & \textbf{SCR}$\uparrow$ & \textbf{TSS}$\uparrow$ & \textbf{PRS}$\uparrow$ \\
  \midrule
  \rowcolor[gray]{0.9}
\multicolumn{8}{l}{\textit{Closed-Source LLMs}} \\
\midrule
Claude3.5-Haiku & 49.60 & 35.05 & 41.07 & 2.30 & 84.20 & 62.74 & 52.12 \\
Claude3.5-Sonnet & 57.55 & 61.71 & 59.55 & 1.52 & 94.52 & 73.20 & 62.68 \\
GPT-3.5 & 46.63 & 51.41 & 48.90 & 1.48 & 93.20 & 67.75 & 51.22 \\
GPT-4 & 59.43 & 63.09 & 61.21 & 1.53 & 93.42 & 71.55 & 61.82 \\
GPT-4o-Mini & 57.95 & 56.43 & 57.18 & 1.67 & 92.98 & 71.82 & 61.52 \\
GPT-4o & \textbf{64.82} & \textbf{74.50} & \textbf{69.33} & {1.33} & 94.52 & 76.94 & 67.65 \\
  \midrule
  \rowcolor[gray]{0.9}
\multicolumn{8}{l}{\textit{Open-Source LLMs}} \\
\midrule
Mistral-7B-Instruct-v0.3 & 26.01 & 34.90 & 29.81 & {1.21} & 92.55 & 51.92 & 29.57 \\
LLaMA3-8B-Instruct & 44.47 & 25.33 & 32.27 & 2.86 & 80.92 & 51.57 & 42.54 \\
LLaMA3-70B-Instruct & 56.82 & 38.80 & 46.11 & 2.38 & 86.38 & 66.56 & 56.40 \\
Qwen2.5-7B-Instruct & 55.50 & 55.30 & 55.40 & 1.64 & 91.43 & 69.32 & 57.53 \\
Qwen2.5-72B-Instruct & 61.90 & {70.36} & {65.86} & 1.36 & 94.10 & 73.99 & 64.15 \\
DeepSeek-V3 & 56.47 & \underline{71.32} & 63.03 & \underline{1.20} & {95.26} & 74.76 & 62.76 \\
  \midrule
  \rowcolor[gray]{0.9}
\multicolumn{8}{l}{\textit{Specialized Models}} \\
\midrule
xLAM-7b-fc-r & 0.27 & 0.54 & 0.36 & \textbf{0.80} & 88.15 & 11.45 & 4.60 \\
gorilla-openfunctions-v2 & 10.11 & 7.13 & 8.36 & 2.31 & 70.83 & 37.90 & 19.23 \\
ToolLLaMA-2-7b-v2 & 1.89 & 1.34 & 1.57 & 2.29 & 58.77 & 18.29 & 5.01 \\
Mistral-Interact & 4.99 & 4.16 & 4.53 & 1.95 & 83.10 & 25.47 & 9.89 \\
  \midrule
  \rowcolor[gray]{0.9}
\multicolumn{8}{l}{\textit{Ours}} \\
\midrule
\textsc{AskToAct}-LoRA-SFT-7B & 57.68 {\scriptsize ($\uparrow$2.18)} & 63.41 {\scriptsize ($\uparrow$8.11)} & 60.41 {\scriptsize ($\uparrow$5.01)} & 1.48 {\scriptsize ($\downarrow$0.16)} & {\underline{96.05}} {\scriptsize ($\uparrow$4.62)} & {\underline{81.42}} {\scriptsize ($\uparrow$12.10)} & {\underline{68.71}} {\scriptsize ($\uparrow$11.18)} \\
\textsc{AskToAct}-Full-SFT-7B & {\underline{63.88}} {\scriptsize ($\uparrow$8.38)} & {68.10} {\scriptsize ($\uparrow$12.80)} & {\underline{65.92}} {\scriptsize ($\uparrow$10.52)} & 1.53 {\scriptsize ($\downarrow$0.11)} & \textbf{97.37} {\scriptsize ($\uparrow$5.94)} & \textbf{84.55} {\scriptsize ($\uparrow$15.23)} & \textbf{73.12} {\scriptsize ($\uparrow$15.59)} \\

\bottomrule
\end{tabular}
}
\caption{Main results.}
\label{tab:result-daily}
\end{table*}
% \renewcommand{\arraystretch}{1.05}
% \begin{table*}[!h]
% \centering
% \small
% \resizebox{0.85\textwidth}{!}{
% \begin{tabular}{lccc}
% \toprule
% \textbf{Metric} & \textbf{Qwen2.5-7B-Instruct} & \textbf{\textsc{AskToAct}-LoRA-SFT-7B} & \textbf{Improvement} \\
% \midrule
% Task Completion Rate (\%)    & 86.67 & {93.33} & +6.66 \\
% Intent Coverage Rate (\%)       & 65.85 & {75.61} & +9.76 \\
% Interaction Rounds  & 3.20  & {2.73}  & -0.47 \\
% User Satisfaction Score (1--5)       & 3.80/5  & {4.40/5}  & +0.60 \\
% \bottomrule
% \end{tabular}
% }
% \caption{Human evaluation results comparing {Qwen2.5-7B-Instruct} and our fine-tuned model {\textsc{AskToAct}-LoRA-SFT-7B}. All metrics are averaged across participants.}
% \label{tab:human_eval}
% \end{table*}

\renewcommand{\arraystretch}{1.05}
\begin{table*}[!h]
\centering
\small
\resizebox{0.95\textwidth}{!}{
\begin{tabular}{lccc}
\toprule
\textbf{Metric} & \textbf{{Qwen2.5-7B-Instruct}} & \textbf{\textsc{AskToAct}-LoRA-SFT-7B} & \textbf{\textsc{AskToAct}-Full-SFT-7B} \\
\midrule
Task Completion Rate (\%)    & 86.67 & 93.33 {\scriptsize ($\uparrow$6.66)} & 96.67 {\scriptsize ($\uparrow$10.00)} \\
Intent Coverage Rate (\%)    & 65.85 & 75.61 {\scriptsize ($\uparrow$9.76)} & 85.37 {\scriptsize ($\uparrow$19.52)} \\
Interaction Rounds           & 3.20  & 2.73 {\scriptsize ($\downarrow$0.47)} & 2.60 {\scriptsize ($\downarrow$0.60)} \\
User Satisfaction Score (1--5) & 3.80 & 4.40 {\scriptsize ($\uparrow$0.60)} & 4.61 {\scriptsize ($\uparrow$0.81)} \\

\bottomrule
\end{tabular}
}
\caption{Human evaluation results.
% comparing {Qwen2.5-7B-Instruct}, \textsc{AskToAct}-LoRA-SFT-7B, and \textsc{AskToAct}-Full-SFT-7B. 
All metrics are averaged across participants.}
\label{tab:human_eval}
\end{table*}

\subsection{Metrics}

We evaluate the models in two aspects: \textit{intent clarification quality} and \textit{tool invocation accuracy}. 
% \paragraph{Intent Clarification Quality}
For intent clarification quality, we design four metrics. \textit{Intent Coverage Rate (ICR)} measures the proportion of successfully clarified intents among all unspecified intents, while \textit{Clarification Efficiency (CE)} evaluates the success rate of clarification across interaction rounds. We combine these measures into a \textit{Clarification Performance Score (CPS)} using a harmonic mean, similar to the F1-score formulation. Additionally, we track \textit{Interaction Rounds (IR)} as the average number of clarification rounds per query.
For tool invocation accuracy, we introduce three complementary metrics. \textit{Solution Completion Rate (SCR)} measures the proportion of successfully generated tool invocation solutions, providing an end-to-end assessment. \textit{Tool Selection Score (TSS)} evaluates API selection accuracy using an F1-score over selected and required APIs. \textit{Parameter Resolution Score (PRS)} assesses the accuracy of parameter resolution through an F1-score computation over API-parameter-value triples.
The details are provided in Appendix~\ref{sec:metrics}.

\subsection{Main Results}

\subsubsection{LLM-based Simulated Evaluation}

% The experimental results are presented in Table~\ref{tab:result-daily}. 
The experimental results on the in-domain (ID) test split of the xlam-IC dataset are presented in Table~\ref{tab:result-daily}.
Our method demonstrates superior performance in both intent clarification and tool invocation.

\paragraph{Intent Clarification Capability}
Both variants of our method—\textsc{AskToAct}-LoRA-SFT-7B and \textsc{AskToAct}-Full-SFT-7B—exhibits strong capabilities in intent clarification. In particular, the fully fine-tuned variant reaches a CPS  of 65.92\%, closely approaching the performance of SOTA LLMs such as GPT-4o.
Meanwhile, the lightweight LoRA variant also achieves competitive results (ICR: 57.68\%, CE: 63.41\%, CPS: 60.41\%), significantly surpassing  the specialized intent clarification model Mistral-Interact.

% These results underscore the effectiveness and robustness of our method, demonstrating consistent performance gains across different fine-tuning settings.

\paragraph{Tool Invocation Accuracy}

Our method demonstrates remarkable capabilities in translating clarified intents into precise tool invocations. It achieves SOTA performance across all evaluation metrics (SCR > 96\%, TSS > 81\%, PRS > 68\%), significantly surpasses all existing open-source and closed-source LLMs. 
Compared to tool-augmented LLMs such as xLAM-7b-fc-r, gorilla-openfunctions-v2, and ToolLLaMA-2-7b-v2, our method demonstrates substantial advantages. This performance gap highlights the strength of integrating intent clarification with tool learning.  Unlike prior specialized models that are limited to unambiguous tool-use queries, our method effectively resolves ambiguity in user queries, leading to significantly improved tool invocation accuracy. 

Further analyses—including cross-model transferability, the impact of augmentation proportion, clarification complexity, and a case study of user interaction styles—are presented in Appendix~\ref{sec:analysis}.

\renewcommand{\arraystretch}{1.05}
\begin{table*}[h]
\centering
\small
\resizebox{\textwidth}{!}{
\begin{tabular}{l ccccccc}
\toprule
\multirow{2}{*}{\centering \textbf{LLM}} & \multicolumn{4}{c}{\textbf{Intent Clarification Quality}} & \multicolumn{3}{c}{\textbf{Tool Invocation Accuracy}} \\
\cmidrule(lr){2-5} \cmidrule(lr){6-8} 
  & \textbf{ICR}$\uparrow$ & \textbf{CE}$\uparrow$ & \textbf{CPS}$\uparrow$ & \textbf{IR}$\downarrow$ & \textbf{SCR}$\uparrow$ & \textbf{TSS}$\uparrow$ & \textbf{PRS}$\uparrow$ \\
  \midrule
  \rowcolor[gray]{0.9}
\multicolumn{8}{l}{\textit{Closed-Source LLMs}} \\
\midrule
Claude3.5-Haiku & {61.07} & 29.88 & 40.13 & 4.20 & 73.68 & 66.15 & 46.59 \\
Claude3.5-Sonnet & 69.74 & 38.10 & 49.28 & 3.29 & 84.96 & 76.05 & 54.09 \\
GPT-3.5 & 44.19 & 44.42 & 44.30 & \textbf{1.72} & 98.27 & 89.28 & 45.50 \\
GPT-4 & 63.60 & 44.73 & 52.52 & 2.57 & 93.90 & 90.52 & 63.26 \\
GPT-4o-mini & \underline{70.86} & 49.60 & 58.35 & 2.63 & 95.22 & 89.63 & {69.44} \\
GPT-4o & \textbf{72.41} & 53.37 & \underline{61.45} & 2.27 & 96.24 & 92.22 & \textbf{69.56} \\
  \midrule
  \rowcolor[gray]{0.9}
\multicolumn{8}{l}{\textit{Open-Source LLMs}} \\
\midrule
Mistral-7B-Instruct-v0.3 & 55.13 & 28.94 & 37.96 & 3.15 & 77.34 & 68.73 & 49.35 \\
LLaMA3-8B-Instruct & 62.81 & 29.99 & 40.59 & 3.66 & 78.86 & 69.22 & 45.52 \\
LLaMA3-70B-Instruct & 67.14 & 35.34 & 46.30 & 3.44 & 84.76 & 79.08 & 55.53 \\
Qwen2.5-7B-Instruct & 64.79 & 38.43 & 48.25 & 3.00 & 92.99 & 86.19 & 61.86 \\
Qwen2.5-72B-Instruct & 68.64 & 43.87 & 53.53 & 2.83 & 92.48 & 90.25 & 63.85 \\
DeepSeek-V3 & 60.11 & 42.24 & 49.62 & 2.55 & 92.17 & 83.10 & 58.33 \\
  \midrule
  \rowcolor[gray]{0.9}
\multicolumn{8}{l}{\textit{Specialized Models}} \\
\midrule
xLAM-7b-fc-r & 0.34 & 0.56 & 0.42 & \underline{2.08} & 91.46 &  14.29 & 5.43\\
gorilla-openfunctions-v2 & 44.53 & 22.27 & 29.69 & 3.41 & 69.92 & 52.36 & 21.67 \\
ToolLLaMA-2-7b-v2 & 2.76 & 2.25 & 2.48 & {2.19} & 98.98 & 42.65 & 2.07 \\
Mistral-Interact & 35.38 & 15.63 & 21.69 & 4.27 & 64.43 & 18.60 & 2.94 \\
  \midrule
  \rowcolor[gray]{0.9}
\multicolumn{8}{l}{\textit{Ours}} \\
\midrule
\textsc{AskToAct}-LoRA-SFT-7B & 68.87 {\scriptsize ($\uparrow$4.08)} & {\underline{53.52}} {\scriptsize ($\uparrow$15.09)} & {60.23} {\scriptsize ($\uparrow$11.98)} & 2.82 {\scriptsize ($\uparrow$0.18)} & {\underline{99.59}} {\scriptsize ($\uparrow$6.63)} & \textbf{96.44} {\scriptsize ($\uparrow$10.25)} & 64.81 {\scriptsize ($\uparrow$2.95)} \\
\textsc{AskToAct}-Full-SFT-7B & 69.90 {\scriptsize ($\uparrow$5.11)} & \textbf{57.27} {\scriptsize ($\uparrow$18.84)} & \textbf{62.96} {\scriptsize ($\uparrow$14.71)} & 2.72 {\scriptsize ($\uparrow$0.28)} & \textbf{99.70} {\scriptsize ($\uparrow$6.64)} & {\underline{96.41}} {\scriptsize ($\uparrow$10.22)} & {\underline{69.45}} {\scriptsize ($\uparrow$7.59)} \\

\bottomrule
\end{tabular}
}
\caption{OOD generalization performance comparison.}
\label{tab:result-ood}
\end{table*}

\begin{table*}[!h]
\centering 
\resizebox{\textwidth}{!}{
\begin{tabular}{lccccccc}
\toprule
\multirow{3}{*}{\centering \textbf{Method}} & \multicolumn{5}{c}{\textbf{Error Rate  (\%)}} & \multirow{3}{*}{\centering \textbf{CPS}$\uparrow$} &\multirow{3}{*}{\centering \textbf{PRS}$\uparrow$} \\
\cmidrule(lr){2-6}
 & \makecell[c]{\textbf{Clearly Stated }\\\textbf{Intent Clarification}} & \makecell[c]{\textbf{Imprecise}\\\textbf{Clarification}} & \makecell[c]{\textbf{Irrelevant}\\\textbf{Clarification}} & \makecell[c]{\textbf{Redundant}\\\textbf{Clarification}} & \makecell[c]{\textbf{Incomplete}\\\textbf{Clarification}} \\
\midrule
\textsc{AskToAct}-LoRA-SFT-7B & 6.80 & 11.65 & 8.74 & 6.80 & 32.00 & 61.51 & 69.50 \\
\quad w/o Error-Correction Augmented Dialogue Data & 9.09 & 6.49 & 6.49 & 9.09 & 38.00 & 58.93 &  66.90\\
\quad w/o Training (Base Model) & 12.43 & 12.37 & 9.29 & 11.34 & 44.00 &  53.00 & 59.18 \\
\bottomrule
\end{tabular}
}
\caption{Ablation study. The first four error types calculated as the proportion of interaction turns containing specific errors among all interaction turns, while the last error type measures the proportion of queries in which not all unspecified intents are successfully clarified.}
\label{tab:ablation}
\end{table*}

\subsubsection{Human-Interactive Evaluation}
To assess the effectiveness of our method in real-world interactions, we conducted a human-interactive evaluation. We recruited 3 participants, each asked to propose 10 unspecified tool-use queries requiring clarification. These queries were independently tested on the base model ({Qwen2.5-7B-Instruct}) and our models ({\textsc{AskToAct}-LoRA-SFT-7B} and {\textsc{AskToAct}-Full-SFT-7B}). Participants interacted with the models iteratively until they obtained a satisfactory response.

As shown in Table~\ref{tab:human_eval}, both variants of our method outperform the base model across all metrics. Specifically, {\textsc{AskToAct}-LoRA-SFT-7B} improves the Task Completion Rate by 6.66\% and the Intent Coverage Rate by 9.76\%, while reducing Interaction Rounds from 3.20 to 2.73. {\textsc{AskToAct}-Full-SFT-7B} achieves further improvements, reaching 96.67\% Task Completion Rate and 85.37\% Intent Coverage Rate, with fewer Interaction Rounds (2.60). In addition, participants reported higher satisfaction with both variants (4.40 and 4.61 vs. 3.80), confirming that our method leads to a more effective and user-friendly interaction experience.
 % our model outperforms the base model across all four evaluation metrics. Specifically, it improves the Task Completion Rate by 6.66\% and the Intent Coverage Rate by +9.76\%. Notably, it reduces the average number of Interaction Rounds from 3.20 to 2.73, demonstrating its ability to resolve unspecified queries more efficiently. 
% In addition, participants reported higher satisfaction with our model (4.40 vs. 3.80), confirming that our model leads to a more effective and user-friendly interaction experience.
% While the LLM-based evaluation simulates user interactions, the human-interactive setting allows us to evaluate the model's behavior in actual multi-turn interactions with real users. The consistency of results across both evaluations reinforces the effectiveness, stability, and practical utility of our proposed method.
The consistency of results across both LLM-based and human-interactive evaluations highlights the effectiveness, robustness, and practical utility of our method.

\subsection{OOD Generalization}

% \subsection{OOD Robustness}
To assess the generalization ability of our method, we test on Taskbench-IC, an out-of-domain (OOD) set that consists of entirely unseen API domains.

As shown in Table~\ref{tab:result-ood}, both the LoRA and fully fine-tuned variants of our method demonstrate strong performance. The LoRA variant achieves a CPS of 60.23\% and PRS of 64.81\%, outperforming all open-source baselines and even surpassing some commercial closed-source models. The fully fine-tuned variant pushes this further, reaching a CPS of 62.96\% and PRS of 69.45\%, comparable to GPT-4o. These results highlight that our method generalizes effectively to unseen domains without relying on memorization of training data. Instead, it acquires transferable principles for intent clarification and tool invocation.

% \subsection{泛化性？？}

% \subsubsection{在ood数据集上的泛化性测试}
% We further evaluate the model's generalization capability on an OOD test set, which comprises API domains that are entirely different from our training data. We conduct experiments on Level III, as this represents the most challenging scenario. 

%  As shown in Table~\ref{tab:result-ood}, our model achieves competitive performance on unseen API domains. Specifically, our model achieves CPS of 51.43\% and  PRS of 52.55\%, which stands  higher than existing open-source models and is comparable to top-tier closed-source models. The strong OOD performance suggests that our model successfully captures generalizable patterns, rather than merely memorizing domain-specific features.

\subsection{Ablation Study}
To assess the contribution of each component in our method,
% To evaluate the effectiveness of \textsc{AskToAct} training,  
we conducted a comprehensive ablation study comparing three model configurations: (1) \textsc{AskToAct}-LoRA-SFT-7B model, (2) a variant without error-correction augmented dialogue data  (i.e., trained only with basic intent clarification data using the same LoRA configurations), and (3) the untrained base model (Qwen2.5-7B-Instruct). We randomly selected 50 unspecified user queries from the test set and computed the error rates for five error types identified in §\ref{sec:error_type}.

As shown in Table~\ref{tab:ablation}, compared to the untrained base model, 
 the model  trained solely on basic intent clarification data significantly reduce all five error types, confirming the effectiveness of  clarification training. Incorporating error-correction augmented dialogues and self-correction training yields further improvements. The Clearly Stated Intent Clarification rate and Redundant Clarification rate both decrease from 9.09\% to 6.80\%, suggesting that the model becomes more effective at avoiding unnecessary clarification. While Imprecise and Irrelevant Clarification rates show slight increases, likely due to additional interaction turns introduced by self-correction attempts, this trade-off is justified by the substantial reduction in Incomplete Clarification rate (from 38.00\% to 32.00\%), which is critical for enabling accurate tool invocation. These improvements in clarification behavior are further reflected in downstream performance.  The PRS increases from 59.18\% to 66.90\%, and finally to 69.50\%, indicating that the enhanced clarification quality translates into more accurate tool invocation.

% Overall, the ablation study confirms that both components contribute meaningfully to reducing generation errors and improving clarification quality in multi-turn interactions. 
% Appendix~\ref{sec:case_study} presents a representative case study comparing the outputs of all three model, providing qualitative insight into the impact of each component.

% As shown in Table~\ref{tab:ablation}, the experimental results demonstrate the effectiveness of self-correction training, significantly enhancing the model's intent clarification capabilities through dynamic error detection and correction. Compared to the model fine-tuned with basic dialogue data, our model achieves further improvements in several aspects. The Clearly Stated Intent  rate and Redundant Clarification  rate both decreases from 9.09\% to 6.80\%. While Imprecise and Irrelevant Clarification rates show slight increases due to additional interaction turns from self-correction attempts, this trade-off is justified by the substantial reduction in Incomplete Clarification rate (from 38.00\% to 32.00\%), indicating more thorough intent understanding.

\section{Related Work}
Our work relates to three areas: \textit{tool learning}, \textit{user intent clarification}, and \textit{self-correction}. Tool learning equips LLMs with external capabilities but typically assumes explicit user intents. Intent clarification addresses ambiguous queries, yet existing datasets often rely on manual annotation. Self-correction has shown promise in mathematical reasoning but remains underexplored for intent understanding. We unify these directions through a self-correcting clarification framework. A full review of related work is provided in Appendix~\ref{sec:related_work}.

% Our work builds on three major research directions: tool learning, user intent clarification, and self-correction mechanisms. First, recent advances in \textit{tool learning} enhance LLMs by enabling access to external APIs or databases, but most evaluations assume user intents are explicitly stated. Second, \textit{user intent clarification} has been explored to handle underspecified queries via interaction, yet existing datasets often rely on costly manual annotation. Third, \textit{self-correction} has shown promise in mathematical reasoning, but remains underexplored for intent understanding in open-ended scenarios. We position our work at the intersection of these threads by developing an automated framework for generating clarification-oriented tool-use data, along with a self-correction mechanism for real-time intent refinement. A more comprehensive review of related literature is provided in Appendix~\ref{sec:related_work}.

\section{Conclusion}

In this work, we presented \textsc{\textbf{AskToAct}}, a self-correcting clarification framework for tool learning that addresses the critical challenges of data scalability and error handling in clarification interactions. Our key contribution lies in leveraging the inherent structure of tool learning datasets to enable automated construction of high-quality intent clarification data, while introducing a novel self-correction mechanism for robust clarification. Experimental results demonstrate that our method not only achieves superior performance in intent clarification and tool invocation but also exhibits strong generalization to unseen APIs and diverse model architectures. We hope that our work will provide valuable insights for developing more effective and reliable intent clarification mechanisms in human-LLM interaction systems.

\section*{Limitations}
While our work demonstrates promising results in handling unspecified queries, several important limitations warrant discussion:

\paragraph{Dataset and Training}
Our method heavily relies on existing tool learning datasets, which may not fully capture the diversity and complexity of real-world user intents. The parameter removal process, although systematic, might not perfectly simulate natural query ambiguity patterns. Additionally, our current approach to error-correction augmentation focuses on pre-defined error types, potentially missing other important error patterns that emerge in real-world interactions.

\paragraph{Interaction Dynamics}
We have not yet explored scenarios where intents must be inferred from previous tool invocation results, limiting our framework's ability to handle context-dependent queries. 
% The current dialogue simulation approach, while structured, may not fully replicate the variability and unpredictability of human responses, particularly in cases where users provide partial or inconsistent information.

\paragraph{Evaluation Limitations}
While our multi-level evaluation framework provides comprehensive assessment, it may not fully capture the complexity of real-world deployment scenarios, particularly in terms of user patience, time constraints, and varying expertise levels. The current metrics might not sufficiently measure the user experience aspects of the clarification process.

\section*{Ethics Statement}

We acknowledge that all authors are informed about and adhere to the ACL Code of Ethics and the Code of Conduct.

\paragraph{Use of AI-Generated Content}
In our research, we utilize LLMs to generate intent clarification dialogues based on existing tool learning datasets. All AI-generated content has been thoroughly verified by the authors to ensure quality and appropriateness. We have implemented rigorous quality control mechanisms to filter out inappropriate or low-quality generations. The paper clearly discloses all instances where AI systems contributed to content generation.

\paragraph{Data Sources}
The tool learning datasets used in our experiments are derived from publicly available sources, including open-source repositories and publicly released benchmarks. We have made reasonable efforts to ensure that these data sources do not contain personally identifiable information  or legally protected content. However, we cannot guarantee that they are entirely free from socially harmful or biased language. Any potential biases in the original datasets may propagate to our results.

\paragraph{Broader Impact}
Our work aims to enhance models’ ability to handle ambiguous user queries in tool-use scenarios. This may extend the applicability of AI systems to a wider range of real-world scenarios. However, such improvements in intent clarification and tool-use capabilities could also enable models to act with limited human oversight, posing both opportunities and risks depending on the deployment context.
% Our work aims to improve human-LLM interaction by reducing the burden on users to formulate precise queries. This has potential positive societal impacts by making AI tools more accessible to a broader population. However, we recognize that improvements in tool-use capabilities could potentially enable automated systems to perform actions with less human oversight, which carries both benefits and risks depending on the application context.

% \paragraph{AI Assistance for Writing}
% We used AI assistants to help with grammar checking and formatting in this paper, but all technical content, analysis, and conclusions represent the original work of the authors.

% Bibliography entries for the entire Anthology, followed by custom entries
%\bibliography{anthology,custom}
% Custom bibliography entries only
\bibliography{custom}

\clearpage

\appendix

\section*{Appendix}
\label{sec:appendix}

\section{Related Work}
\label{sec:related_work}
\subsection{Tool Learning}

Tool learning can effectively alleviate the inherent limitations of LLMs through dynamic interaction with external tools \cite{NEURIPS2023_d842425e,  tang2023toolalpacageneralizedtoollearning,shen2023hugginggpt,qin2023toolllm,wang2024empowering}. While LLMs acquire limited knowledge during the pre-training phase, tools such as integrated search engines \cite{nakano2021webgpt, komeili-etal-2022-internet, NEURIPS2023_d842425e, zhang2023toolcoderteachcodegeneration, shi2023replugretrievalaugmentedblackboxlanguage, paranjape2023artautomaticmultistepreasoning} and databases \cite{thoppilan2022lamdalanguagemodelsdialog, patil2023gorilla,NEURIPS2023_8fd1a81c, NEURIPS2023_9cb2a749, chen2023tooldec, gu-etal-2024-middleware} enable real-time access to up-to-date information beyond the training data. In addition, LLMs often struggle with complex mathematical operations, code generation, and domain-specific tasks \cite{inaba-etal-2023-multitool,  bran2023augmenting}, which can be enhanced through dedicated tools.

Existing evaluation benchmarks for reliable tool usage \cite{huang2023metatool, patil2023gorilla, ning2024wtuevalwhetherornottoolusage} focus on explicit and unambiguous user queries, leaving the challenges of handling unspecified intents in real-world scenarios largely unexplored.

\subsection{User Intent Clarification}
When interacting with users, understanding user intents is crucial, especially when intents are implicit or unspecified. \citet{zhang2023clarify} shows that unspecified user intents in queries should be clarified through interaction.  The STaR-GATE framework \cite{andukuri2024star} introduces a systematic approach to question formulation by simulating diverse clarification scenarios. \citet{qian-etal-2024-tell} applied several strategies in conversation record construction and leveraged the generated data to fine-tune the model, enhancing the ability to formulate targeted questions.

However, the construction of high-quality datasets for training and evaluation still remains challenging. \citet{qian-etal-2024-tell} constructed a benchmark for daily scenarios, while \citet{wang2024learning} focuses on tool learning scenarios, but they both relied on manual annotation. Our work introduces an automated pipeline for dataset construction, enabling better scalability.

\subsection{Self-Correction Mechanism}
Early work on self-correction \cite{huang-etal-2023-large, madaan2023selfrefine} primarily focused on post-correction, using feedback to improve model outputs after they are generated. However,  \citet{huang2024large} found that in the absence of standardized answers, such post-correction has limited effect. This finding prompted a shift in research focus to real-time self-correction, i.e., dynamically identifying and correcting errors during the reasoning process. 

Self-correction has achieved success in mathematical reasoning, where \citet{yan2024s3cmathspontaneoussteplevelselfcorrection} and \citet{zhang2024learningcheckunleashingpotentials} introduce step-level and multi-granular correction strategies. We extend these approaches to user intent clarification, enabling real-time correction during the clarification process.

\section{Datasets}
\label{sec:dataset}
Our dataset is constructed based on two existing tool learning datasets: xlam-function-calling-60k and TaskBench. We describe their characteristics below.

\paragraph{xlam-function-calling-60k} This dataset comprises functionally executable APIs extracted from Python libraries and RESTful services, rather than being manually defined. The APIs span 21 functional categories, covering a broad range of domains such as information retrieval,  and computational tools. In total, the dataset contains 3,673 APIs and 60,000 samples.
An example is shown below:
\begin{lstlisting}[style=promptstyle]
{
    "query": "List titles originally aired on networks '1' and '8', released after 2010, sorted by release date in descending order.",
    "tools": [
        {
            "name": "list_titles",
            "description": "Fetches a listing of titles that match specified parameters from the Watchmode API.",
            "parameters": {
                "genres": {
                    "description": "Filter results to only include certain genre(s). Pass in a single genre ID or multiple comma-separated IDs. Default is '4,9'.",
                    "type": "str",
                    "default": "4,9"
                },
                "limit": {
                    "description": "Set how many titles to return per page. Default and maximum is 250.",
                    "type": "int",
                    "default": "250"
                },
                "source_ids": {
                    "description": "Filter the results to titles available on specific sources by passing individual IDs or multiple comma-separated IDs. Default is '23,206'. Note: Only a single region can be set if this is populated.",
                    "type": "str",
                    "default": "23,206"
                },
                "source_types": {
                    "description": "Filter results to only include titles available on specific types of sources (e.g., subscription, free). Default is 'sub,free'. Note: Only a single region can be set if this is populated.",
                    "type": "str",
                    "default": "sub,free"
                },
                "types": {
                    "description": "Filter results to only include titles available on specific types of sources (e.g., subscription, free). Default is 'sub,free'. Note: Only a single region can be set if this is populated.",
                    "type": "str",
                    "default": "movie,tv_series"
                },
                "regions": {
                    "description": "Filter results to only include sources active in specific regions. Currently supported regions: US, GB, CA, AU. Default is 'US'. Note: Only a single region can be set if source_ids or source_types are populated.",
                    "type": "str",
                    "default": "US"
                },
                "sort_by": {
                    "description": "Sort order of results. Possible values include: relevance_desc, relevance_asc, popularity_desc, popularity_asc, release_date_desc, release_date_asc, title_desc, title_asc. Default is 'relevance_desc'.",
                    "type": "str",
                    "default": "relevance_desc"
                },
                "page": {
                    "description": "Set the page of results to return. Default is 1.",
                    "type": "int",
                    "default": "1"
                },
                "network_ids": {
                    "description": "Filter results to titles that originally aired on specific TV networks by passing individual IDs or multiple comma-separated IDs. Default is '1,8,12'.",
                    "type": "str",
                    "default": "1,8,12"
                },
                "release_date_start": {
                    "description": "Filter results to only include titles released on or after a specific date. Format: YYYYMMDD. Default is 20010101.",
                    "type": "int",
                    "default": "20010101"
                },
                "release_date_end": {
                    "description": "Filter results to only include titles released on or before a specific date. Format: YYYYMMDD. Default is 20201211.",
                    "type": "int",
                    "default": "20201211"
                }
            }
        }
    ],
    "answers": [
        {
            "task": "list_titles",
            "arguments": [
                { "name": "network_ids", "value": "1,8" },
                { "name": "release_date_start", "value": 20110101 },
                { "name": "sort_by", "value": "release_date_desc" }
            ]
        }
    ]
}
\end{lstlisting} 

\paragraph{Taskbench} In contrast, {TaskBench} defines three tool-use scenarios across distinct application domains:
(1) {Hugging Face tools}, simulating a subset of model functionalities available on the Hugging Face platform—such as summarization, translation, and classification, with 23 APIs and 12,217  samples;
(2) {Multimedia tools}, simulating representative functionalities of multimodal systems—such as video editing and image transformation, with 40 APIs and 8,904 samples;
(3) {Daily Life APIs}, simulating everyday user-facing applications—such as ticket booking, food ordering and schedule management, with 40 APIs and 7,150 samples.
All APIs in {TaskBench} are manually constructed. Representative examples from each domain are shown below: 

\noindent\rule{\linewidth}{0.5pt}
\textit{Hugging Face tools}  

\begin{lstlisting}[style=promptstyle]
{
    "query": "I'm currently analyzing a particular text, 'John works at Google in Mountain View, California.' Can you assist me in identifying the named entities and marking the part-of-speech tags within this text?",
    "tools": [
        {
            "id": "Token Classification",
            "desc": "Token classification is a natural language understanding task in which a label is assigned to some tokens in a text. Some popular token classification subtasks are Named Entity Recognition (NER) and Part-of-Speech (PoS) tagging. NER models could be trained to identify specific entities in a text, such as dates, individuals and places; and PoS tagging would identify, for example, which words in a text are verbs, nouns, and punctuation marks.",
            "input-type": [
                "text or text file"
            ],
            "output-type": [
                "text or text file"
            ]
        }
    ],
    "answers": [
        {
            "task": "Token Classification",
            "arguments": [
                "'John works at Google in Mountain View, California.'"
            ]
        }
    ]
}
\end{lstlisting} 

\noindent\rule{\linewidth}{0.5pt}
\textit{Multimedia tools}  

% 提供两个数据集的例子，说明不同
\begin{lstlisting}[style=promptstyle]
{
    "query": "I've recently conducted an interview and have recorded it in 'interview.wav' audio file. Can you assist me in transcribing it to a text document, so I can refer to it easily in the future? Besides, I'm dealing with an article titled 'abc.txt' and I want to have a fresh iteration of this text so that it will be unique. Would you be able to employ the Article Spinner tool to facilitate this?",
    "tools": [
        {
            "id": "Audio-to-Text",
            "desc": "Transcribes speech from an audio file into text.",
            "input-type": [
                "audio or audio file"
            ],
            "output-type": [
                "text or text file"
            ]
        },
        {
            "id": "Article Spinner",
            "desc": "Rewrites a given article using synonyms and syntax changes to create a new, unique version.",
            "input-type": [
                "text or text file"
            ],
            "output-type": [
                "text or text file"
            ]
        }
    ],
    "answers": [
        {
            "task": "Audio-to-Text",
            "arguments": [
                "interview.wav"
            ]
        },
        {
            "task": "Article Spinner",
            "arguments": [
                "abc.txt"
            ]
        }
    ]
}
\end{lstlisting} 

\noindent\rule{\linewidth}{0.5pt}
\textit{Daily Life APIs}  

\begin{lstlisting}[style=promptstyle]
{
    "query": "I have a busy day ahead. Could you assist me by logging into an online meeting regarding 'Smart Home Devices'? After the meeting, can you facilitate a video call with my friend on +1234567666?",
    "tools": [
        {
            "id": "attend_meeting_online",
            "desc": "Attend a meeting online about a specific topic",
            "parameters": [
                {
                    "name": "topic",
                    "type": "string",
                    "desc": "The topic of the meeting"
                }
            ]
        },
        {
            "id": "make_video_call",
            "desc": "Make a video call to a specific phone number",
            "parameters": [
                {
                    "name": "phone_number",
                    "type": "string",
                    "desc": "The phone number to make the video call to"
                }
            ]
        }
    ],
    "answers": [
        {
            "task": "attend_meeting_online",
            "arguments": [
                { "name": "topic", "value": "Smart Home Devices" }
            ]
        },
        {
            "task": "make_video_call",
            "arguments": [
                { "name": "phone_number", "value": "+1234567666" }
            ]
        }
    ]
}
\end{lstlisting} 

We verified that there is no overlap between APIs in {xlam-function-calling-60k}  and those in {TaskBench}.

\paragraph{Licensing} Both datasets are publicly accessible: {xlam-function-calling-60k} follows the Creative Commons Attribution 4.0 License (CC BY) while {TaskBench} is released under the Apache 2.0 License. We comply with their respective licenses in using and extending the data.

\section{Intent Clarification Dataset Curation}

\subsection{Unspecified Query Generation}

\subsubsection{Prompt for Unspecified Query Generation}
\label{sec:unspecified_query_generation}
Given an original query and its tool invocation solution, the following  prompt guides GPT-4o to generate unspecified queries by analyzing parameters and systematically removing them.

\noindent\rule{\linewidth}{0.5pt}
\textit{System Prompt}  
\begin{lstlisting}[style=promptstyle]
You are a query transformation assistant. Your task is to modify the original user query by removing or abstracting specific parameters marked with `removed`: "true", while maintaining the overall structure and clarity of the original query. The resulting query (`unspecified_query`) should reflect the general intent of the user but omit or obscure the specific details of the removed parameters.

### Input:
1. `original_query`: The complete textual description of the user's original request.
2. `answers`: A detailed record specifying the APIs and parameters required to fulfill the original query. Each parameter in this record includes:
    - `removed`: A boolean ("true" or "false") indicating whether this parameter should be removed or abstracted during the transformation process.
    - Other relevant metadata, such as the parameter's value.
3. `tools`: Documentation or descriptive details about the tools referenced in the query, including their parameters and usage instructions.

### Transformation Rules:
1. Identify the parameters to be removed or abstracted:
    - Focus on parameters where `removed`: "true".
    - Identify the full range of corresponding expressions in the query, ensuring all references to the parameter are appropriately handled.
2. Apply the appropriate transformation strategy to each parameter marked as `removed`: "true":
    - Complete Removal: The parameter is entirely removed when it has no significant impact on the remaining content of the query. However, this should not be used if the parameter is optional. Also, if the same tool is called multiple times, the parameters should not be removed. Instead, they should be abstracted.
    - Semantic Abstraction: If the parameter influences the meaning or structure of the query, replace its value with a more general or abstract term. 
    - Partial Obfuscation: If the elements of a matrix or list are presented separately in the query (e.g., discrete values like quantities or items) and need to be constructed or inferred, only one element from the matirx or list should be removed or abstracted. The remaining elements should stay intact. This can still be done using either complete removal or semantic abstraction, while leaving other relevant elements unchanged.
3. Ensure textual and structural coherence:
    - After transformation, ensure that the `unspecified_query` remains readable, logically consistent, and grammatically correct.
4. Avoid explicitly stating "unspecified" or "unknown" values:
    - Do not use terms like "unspecified", "unknown", or "ambiguous" in the `unspecified_query`.
    - Instead, naturally omit or generalize the missing details without drawing attention to their absence.
5. Retain the rest of the query:
    - Leave unchanged the parts of the query that are not marked for removal, maintaining consistency in format and information.
    
### Output:
Return a JSON object containing:
- `unspecified_query`: The transformed query string with removed/abstracted parameters.
- `key_info`: A JSON array (or object) documenting all parameters, containing the following fields:
    - `original_value`: The expression of the parameter as it appears in the 'original_query' (not the value in the 'answers').
    - `current_value`: The transformed value of the parameter in the `unspecified_query`.
    - `removed`: The boolean flag indicating whether the parameter was removed.
\end{lstlisting}

\subsubsection{Transformation Record Format}
\label{sec:unspecified_json}

For each unspecified query, we maintain a transformation record in the following JSON structure:
{\small
\begin{verbatim}
{
    "original_query": string,
    "unspecified_query": string,
    "key_info": [
        {
            "name": [API_name],
            "arguments": {
                [param_name]: {
                    "original_value": string,
                    "current_value": string,
                    "removed": boolean
                },
                ...
            }
        },
        ...
    }
}
\end{verbatim} 
}

\subsubsection{Human Verification}
\label{sec:human_verification_1}
\renewcommand{\arraystretch}{1.05}
\begin{table*}[!h]
\centering
\small
\resizebox{0.9\textwidth}{!}{
\begin{tabular}{lcccccc}
\toprule
\textbf{Metric} &
\textbf{Naturalness$\uparrow$} & 
\textbf{Consistency$\uparrow$} & 
% \textbf{Semantic Validity$\uparrow$} & 
\textbf{Necessity$\uparrow$} & 
\textbf{Complexity$\uparrow$} & 
\textbf{Diversity$\uparrow$} & 
\textbf{Acceptance Rate (\%)$\uparrow$}  \\
% \textbf{Inter-annotator Agreement (Fleiss' $\kappa$)$\uparrow$} 
\midrule
  {Score} &  4.61/5 & 4.80/5 & 4.03/5 & 3.87/5 & 4.54/5 & 95.92  \\   % & 0.81 
\bottomrule
\end{tabular}
}
\caption{Human verification results of unspecified query generation. The first five metrics are rated on a 1–5 scale, while Acceptance Rate is reported as a percentage. All metrics are averaged across participants.}
\label{tab:quality_control1}
\end{table*}

% \vspace{1mm}
% \noindent \textbf{Inter-annotator Agreement:} Fleiss' $\kappa$ = 0.81

To further ensure the quality of generated unspecified queries, we perform human verification on 400 randomly sampled queries. Three graduate students with NLP backgrounds independently assessed each query based on six criteria: Naturalness (fluency and linguistic coherence), Consistency (uniformity of transformation), Necessity (need for clarification), Complexity (difficulty of clarification), Diversity (range of parameter types and domains), and Acceptance Rate (overall acceptability). Results are shown in Table~\ref{tab:quality_control1}.

\subsection{Clarification Dialogue Construction}
\label{sec:clarification_generation}

We divide the clarification dialogue construction process into two steps: GPT-4o-dependent content generation and template-based dialogue assembly. The essential content is generated using GPT-4o and stored a structured transformation record (see Appendix~\ref{sec:unspecified_json} for format details). The information encoded in this record is then used to deterministically assemble the final dialogue through predefined templates, without further reliance on GPT-4o.

\subsubsection{GPT-4o-Dependent Content Generation}
\label{sec:json}

\paragraph{Task Decomposition}
   We employ the following prompt to guide GPT-4o in decomposing user queries into subtasks:

\noindent\rule{\linewidth}{0.5pt}
\textit{System Prompt}  
\begin{lstlisting}[style=promptstyle]
You are a smart task decomposition assistant. Your goal is to break down the user's main task into smaller, manageable subtasks. Please follow the instructions below.

You will receive a JSON-formatted input containing:
- `query`: A description of the main task the user wants to accomplish.
- `tools`: A list of APIs available to solve the task, each with a unique identifier and a description of its functionality. Note: The APIs are provided in the exact order necessary to resolve the task.

### Task Decomposition:
1. Analyze the query to understand the user's main task.
2. Break it down into smaller, manageable subtasks that can be handled using the provided APIs. Ensure that each subtask is completed by calling one of the APIs in the exact order they are listed.

Your output should be a JSON object with the following structure:
{
    "tool_steps": [
        "Step <number>: <subtask description> using <API name>.",
        ...
    ]
}
\end{lstlisting}

The decomposition result is added to the transformation record as a new field "tool\_steps".

\paragraph{Clarification Question Generation}
We use the following prompt to generate clarification questions for unspecified parameters:

\noindent\rule{\linewidth}{0.5pt}
\textit{System Prompt}  
\begin{lstlisting}[style=promptstyle]
You are an assistant responsible for generating clarification questions for missing information in the user's query.

### Input:
The input should contain the following fields:
- `original_query`: A complete user task description.
- `unspecified_query`: A user task description missing some key information.
- `tools`: Documentation or descriptive details about the tools referenced in the query, including their parameters and usage instructions.
- `key_info`: This should record the APIs and parameters needed to solve the user task, including information about any missing parameters.
    - `original_value`: The original value of the parameter in the `original_query`.
    - `current_value`: The current value of the parameter in the `unspecified_query`.
    - `removed`: Indicates if the parameter's value is clear ("false") or unspecified ("true").
    
### Task Requirements:
For each parameter where the field `removed` is set to true, you are to generate a clarification question.
- If multiple API calls rely on the same missing information, form a single combined question to efficiently gather the required details, rather than asking multiple separate questions.
- Each question should focus on gathering one specific piece of information to improve the precision of the query and avoid ambiguity.
- Do not ask about information that can be inferred from context or API interactions. Only generate clarification questions for details that cannot be deduced from the given context or API responses.
- Add a `question` field to the corresponding parameter in `key_info`, which contains the generated clarification question.
- Do not modify the `original_query`, `current_value` or any other fields in `key_info`.

### Output:
Only output the modified `key_info` in JSON format, ensuring that the question field contains the clarification question for each missing parameter.
\end{lstlisting}

The newly generated "key\_info" field replaces the original one in the transformation record.

\subsubsection{Template-based Dialogue Assembly}
\label{sec:template-based-generation}
Based on the completed transformation record, we automatically construct the dialogue through predefined templates.

\paragraph{Task Decomposition}
We concatenate steps from "tool\_steps" to form a comprehensive task decomposition analysis.

\paragraph{Parameter Evaluation}
   For each parameter in the transformation record, we generate evaluation statements using templates based on their removed status:
\begin{itemize}
\item For parameters clearly stated in the query, we generate the evaluation that "The parameter [param\_name] for API [API\_name] has a value of [value]". 
\item For parameters removed in the query, we generate the evaluation that  "The parameter [param\_name] for API [API\_name] lacks a clear value".
\end{itemize}

\paragraph{Clarification Interaction}
Following the API call order, for each removed parameter, we generate a three-part clarification interaction:
\begin{itemize}
\item Assistant $\rightarrow$ User: Ask the clarification question.
\item User $\rightarrow$ Assistant: Provide the original parameter value using templates from Table~\ref{tab:user-template}.
\item Assistant: Confirm with "Now I know that the parameter [param\_name] for API [API\_name] has a value of [value]".
\end{itemize}

\begin{table*}[t]
\centering \small
\resizebox{0.8\textwidth}{!}{
\begin{tabular}{>{\centering\arraybackslash}p{1.7cm} 
                >{\raggedright\arraybackslash}p{9.5cm} }
% \begin{tabular}{c l}
% \midrule\noalign{\hrule height 1.5pt}
\toprule
\multicolumn{1}{l}{\textbf{Tone}} & \multicolumn{1}{l}{\textbf{Template}} \\
% \noalign{\hrule height 1.5pt}
\midrule
\multicolumn{1}{l}{\multirow{3}{*}{Neutral}} & [value]. \\
& The answer is: [value]. \\
& Ah, the answer is simply [value]. \\
\midrule
\multicolumn{1}{l}{\multirow{2}{*}{Friendly}} & Sure! The answer is [value]. Let me know if you have more questions! \\
& I'm glad to help! The answer is absolutely [value]! \\
\midrule
\multicolumn{1}{l}{\multirow{3}{*}{Dismissive}} & Honestly, I don't see why this is a big deal, but the answer is [value]. \\
& Okay, the answer is: [value]. Hope that helps, I guess. \\
& Whatever. The answer is [value]. Not that it matters. \\

\midrule
\multicolumn{1}{l}{\multirow{6}{*}{Irritated}} 
& Listen, the answer is [value]. Just deal with it! \\

& Ugh, seriously? The answer is [value]. Can we move on already? \\
& Honestly, do you really need me to repeat this? The answer is [value]. I can't believe we're still discussing this! It's infuriating! \\
& Enough already! The answer is [value]. Can we please get to the point? I'm tired of this nonsense! It's frustrating! Let's just move on! \\

% \midrule\noalign{\hrule height 1.5pt}
\bottomrule
\end{tabular}
}
\caption{Response templates for user  with varying tones.}
\label{tab:user-template}
\end{table*}

\paragraph{Tool Invocation}
% We  organize and output the final tool invocation solution based on the information in the transformation record (including API call sequence and corresponding parameters).

% We organize and output the final tool invocation solution based on the information stored in the transformation record. Specifically, we follow the same API call order as recorded in "key\_info" field and extract the corresponding parameter values for each API. 
We construct the final tool invocation solution using the "{key\_info}" field from the transformation record, which specifies the sequence of API calls and their associated parameters.
The final output is serialized into the following format:

{\small
\begin{verbatim}
[
    {
        "task": [API_name],                  
        "arguments": [                          
            {
                "name": [param_name],                    
                "value": string | number | boolean
            },
            ...
        ]
    },
    ...
]
\end{verbatim}
}

\paragraph{Final Assembly}

We assemble the complete assistant-user dialogue by sequentially integrating the natural language outputs generated in the previous steps.
We begin with a user message that presents the task description and relevant APIs. The assistant’s response is then constructed by combining the task decomposition and parameter evaluation. For each missing parameter, we insert a three-part clarification interaction comprising the assistant’s  question, the user’s response, and the assistant’s confirmation.
This process is repeated until all missing parameters have been clarified. The dialogue concludes with the assistant presenting the full tool invocation solution.

\section{Self-correction Training}
\subsection{Error Generation}

\label{sec:error_generate}
In our template-based dialogue assembly process (Appendix~\ref{sec:template-based-generation}), the sequence of APIs and their required parameters, as recorded in the "{key\_info}" field of the transformation record, implicitly defines the structure of the final dialogue.
% By examining the assembly process described in Appendix~\ref{sec:template-based-generation}, we observe that the sequence of APIs and their required parameters recorded in the "key\_info" field of the transformation record implicitly reflects the structure of the final dialogue.  
This insight motivates our error generation strategy. For each selected error type, we first identify a parameter  position in the transformation record where the error will be introduced. We then generate the corresponding erroneous behavior and annotate the selected parameter with an {"error"} field to indicate its error type.

We now describe the generation strategies for each of the five  error types in detail.

\paragraph{Clearly Stated Intent Clarification} The prompt for generating instances of questioning clearly stated intent  is designed as follows:

\noindent\rule{\linewidth}{0.5pt}
\textit{System Prompt}  
\begin{lstlisting}[style=promptstyle]
You are a smart assistant. Your task is to generate a JSON object based on the given input. Please follow these instructions:

### Input:
The input should contain the following fields:
- `original_query`: A complete user task description.
- `unspecified_query`: A user task description missing some key information.
- `tools`: Documentation or descriptive details about the tools referenced in the query, including their parameters and usage instructions.
- `key_info`: This should record the APIs and parameters needed to solve the user task, including information about any missing parameters.

### Key Requirements:
1. From the `key_info`, select the {selected_param_index} parameter where `removed` is false and assume that its value is missing.
2. Generate a specific clarification question related to the missing parameter, such that the answer would provide the value from the `original_value` field of that parameter, and save it in the `question` field of that parameter.
3. Set `error`: "type 1" to the modified parameter.
4. No other content in `key_info` should be modified.

### Output:
Only output the modified `key_info` in JSON format, ensuring that the `question` field contains the clarification question.
\end{lstlisting}

\paragraph{Imprecise Clarification}  The prompt for generating imprecise clarification questions is designed as follows:

\noindent\rule{\linewidth}{0.5pt}
\textit{System Prompt}  
\begin{lstlisting}[style=promptstyle]
You are a smart assistant. Your task is to generate a JSON object based on the given input. Please follow these instructions:

### Input:
The input should contain the following fields:
- `original_query`: A complete user task description.
- `unspecified_query`: A user task description missing some key information.
- `tools`: Documentation or descriptive details about the tools referenced in the query, including their parameters and usage instructions.
- `key_info`: This should record the APIs and parameters needed to solve the user task, including information about any missing parameters.

### Key Requirements:
1. From the `key_info`, select the {selected_param_index} parameter where the field `removed` is true and assume that its value is missing.
2. Generate an imprecise clarification question about the missing parameter:
    - This question should seem relevant to the user task.
    - However, it should be less precise than the original question provided in the `question` field of the selected parameter.
    - The goal is to make the question introduce ambiguity, meaning it should be unclear what exactly needs to be answered, thus creating confusion about how to provide a precise and accurate response.
3. Directly add this imprecise question to the selected parameter in the `imprecise_question` field.
4. Set `error`: "type 2" to the modified parameter.
5. No other content in `key_info` should be modified.

### Output:
Only output the modified `key_info` in JSON format, ensuring that the selected parameter now contains the imprecise question.
\end{lstlisting}
\paragraph{Irrelevant Clarification}   The prompt for generating irrelevant clarification questions is designed as follows:

\noindent\rule{\linewidth}{0.5pt}
\textit{System Prompt}  
\begin{lstlisting}[style=promptstyle]
You are a smart assistant. Your task is to generate a JSON object based on the given input. Please follow these instructions:

### Input:
The input should contain the following fields:
- `original_query`: A complete user task description.
- `unspecified_query`: A user task description missing some key information.
- `tools`: Documentation or descriptive details about the tools referenced in the query, including their parameters and usage instructions.
- `key_info`: This should record the APIs and parameters needed to solve the user task, including information about any missing parameters.

### Key Requirements:
1. From the `key_info`, select the {selected_param_index} parameter you encounter.
2. Generate a question that appears relevant to the user task but is actually unhelpful for solving the task using the APIs in `key_info`.
3. Directly add this irrelevant question to the selected parameter in the `irrelevant_question` field.
4. Set `error`: "type 3" to the modified parameter.
5. No other content in `key_info` should be modified.

### Output:
Only output the modified `key_info` in JSON format, ensuring that the selected parameter now contains the irrelevant question.
\end{lstlisting}

\paragraph{Redundant Clarification}   We employed the following algorithm to generate redundant  clarification questions:

\vspace{10pt}
\noindent
\begin{minipage}{\linewidth}
\hrule height 1pt
\vspace{4pt}
\textbf{Algorithm 1} Redundant Clarification Generation
\vspace{4pt}
\hrule height 0.5pt
\begin{flushleft}
1: Input: transformation record $R$ \\

2: $p_{target} \leftarrow \text{Random}({p \in R.\text{params} \mid p.\text{pos} > 0})$ \\
3:  $p_{prev} \leftarrow \text{Random}(p \in R.\text{params} \mid$ \\
$p.\text{pos} < p_{target}.\text{pos} \wedge p.\text{removed} = \text{true})$ \\
4: $q_r \leftarrow p_{prev}.\text{question}$  \\
5: $p_{target}.\text{error} \leftarrow \text{"type 4"}$ \\
6: Add $q_r$  to $p_{target} as a redundant question$ \\
7: \textbf{return} updated transformation record $R'$
\end{flushleft}
\hrule height 1pt
\end{minipage} 

% 可以在每行前面加点注释（xxx:xxx）

% \input{tables/algotithm-redundant} 

\paragraph{Incomplete Clarification}  We employed the following algorithm to generate incomplete  clarification process:

\vspace{10pt}
\noindent
\begin{minipage}{\linewidth}
\hrule height 1pt
\vspace{4pt}
\noindent\textbf{Algorithm 2} Incomplete Clarification Generation
\vspace{4pt}
\hrule height 0.5pt
\begin{flushleft}
1: Input: transformation record $R$ \\
2: $k \leftarrow \text{Random}({i \mid 0 \leq i < |R.\text{params}|})$ \\
3: $P_{known} \leftarrow \{p \in R.\text{params} \mid p.\text{pos} < k\}$ \\
4: $template \leftarrow \text{"<unknown\_*>"}$ \\
5: $S_{tools} \leftarrow \{\}$ \\
6: \textbf{for each} $p \in R.\text{params}$ \textbf{do} \\
7: \quad \textbf{if} $p \notin P_{known}$ \textbf{then} \\
% 7: \quad \textbf{if} $p \notin P_{known}$ \t、extbf{then} \
8: \quad\quad $S_{tools}[p] \leftarrow \text{GenUnkVal}(template, p)$  \\
9: \quad \textbf{else} \\
10: \quad\quad $S_{tools}[p] \leftarrow p.\text{original}$ \\
11: $p_{k}.\text{error} \leftarrow \text{"type 5"}$ \\
12: Add $S_{tools}$ to $p_k$ as an incomplete clarification error  \\
13: \textbf{return} updated transformation record $R'$ \quad \\
\end{flushleft}
\hrule height 1pt
\end{minipage} 

\subsection{Error-Correction Augmentation Dialogue Assembly}
\begin{table*}[t]
\centering
\small
\renewcommand{\arraystretch}{1.25}
\resizebox{0.88\textwidth}{!}{
\begin{tabular}{
    >{\raggedright\arraybackslash}m{3.1cm} 
    >{\raggedright\arraybackslash}m{9.8cm}
}
\toprule
 \textbf{Error Type} &  {\textbf{Correction Template}} \\
\midrule
{ Clearly Stated Intent Clarification (Type 1)} & Sorry, I made a mistake. The parameter [param\_name] for API [API\_name] has a value of [value], which was clearly stated in the user's task. Therefore, I don't need to ask any further questions. \\
\midrule
{Imprecise Clarification  (Type 2)} & Sorry, I made a mistake. My question wasn't clear, so let me rephrase it for better understanding. \\
\midrule
{Irrelevant Clarification  (Type 3)} & Sorry, I made a mistake. This question is not related to using the APIs to resolve the user's task, so I will ask something else. \\
\midrule
{Redundant Clarification (Type 4)} & Sorry, I made a mistake. The parameter [param\_name] for API [API\_name] has a value of [value], which has already been asked in the past, so there is no need to ask again. \\
\midrule
{Incomplete Clarification (Type 5)} & Sorry, I made a mistake. I still lack some key information, so I need to ask further questions. \\
\bottomrule
\end{tabular}
}
\caption{Correction templates for each error type.}
\label{tab:correction-template}
\end{table*}

We follow the same assembly procedure as described in Appendix~\ref{sec:template-based-generation}. The only difference is that, when an {"error"} field is detected in the transformation record, we insert the erroneous behavior into the assistant message at the corresponding dialogue position. We then generate the assistant's correction using the type-specific template defined in Table~\ref{tab:correction-template}, and naturally continue the interaction from that point.

% Based on completed transformation record, we can conveniently inject errors at specific positions. During dialogue generation, we follow the same generation process described in §~\ref{sec:template-based-generation}, but when encountering an injected error, we use error-specific templates to create corresponding corrections, thereby producing dialogue data enhanced with error-correction pairs.

% \input{tables/algotithm-redundant}

\subsection{Human Verification}
\label{sec:human_verification_2}
\renewcommand{\arraystretch}{1.05}
\begin{table*}[t]
\centering
\small
\resizebox{0.75\textwidth}{!}{
\begin{tabular}{lccccc}
\toprule
\textbf{Metric} &
\textbf{Naturalness$\uparrow$} & 
\textbf{Plausibility$\uparrow$} & 
% \textbf{Semantic Validity$\uparrow$} & 
\textbf{Usefulness$\uparrow$} & 
% \textbf{Complexity$\uparrow$} & 
\textbf{Diversity$\uparrow$} & 
\textbf{Acceptance Rate (\%)$\uparrow$}  \\
% \textbf{Inter-annotator Agreement (Fleiss' $\kappa$)$\uparrow$} 
\midrule
  {Score} &  4.57/5 & 3.91/5 & 4.11/5 & 4.28/5 & 91.83\%  \\   % & 0.81 
\bottomrule
\end{tabular}
}
\caption{Human verification results of error-correction augmentation. The first four metrics are rated on a 1–5 scale, while Acceptance Rate is reported as a percentage. All metrics are averaged across  participants.}
\label{tab:quality_control2}
\end{table*}

% \vspace{1mm}
% \noindent \textbf{Inter-annotator Agreement:} Fleiss' $\kappa$ = 0.81

To ensure the validity  and reliability of our error-correction augmentation method, we perform human verification on 200 randomly sampled augmented dialogues. Three graduate students with NLP backgrounds independently assessed each dialogue based on five criteria: Naturalness (fluency and linguistic coherence), Plausibility (realism of the injected error-correction pair),  Usefulness (utility of the error-correction pair for intent clarification),  Diversity (variation in error manifestation and expression styles), and Acceptance Rate (overall acceptability). Results are shown in Table~\ref{tab:quality_control2}.

\subsection{Examples for Error-Correction Augmentation}

\label{sec:reflection_template}

The following examples shows different types of errors and their corresponding corrections. 

\noindent\rule{\linewidth}{0.5pt}
\textit{Example 1: Clearly Stated Intent Clarification}  

% \begin{minipage}{\linewidth}
\begin{lstlisting}[style=promptstyle]
(*@\noindent{\textbf{User:}}@*) The task is: Fetch the latest media of the soccer player with player ID 170085. Some relevant APIs: [{'name': 'player_latest_media', 'description': 'Fetch the latest media for a given player using their player ID.', 'parameters': {'player_id': {'description': 'The unique identifier of the player for whom the latest media is to be fetched.', 'type': 'int', 'default': '12994'}}}, {'name': 'teammedia', 'description': 'Fetches media information for a specific baseball team using the given team ID.', 'parameters': {'is_id': {'description': 'The team ID for which to retrieve the media.', 'type': 'int', 'default': 3633}}}]

(*@\noindent{\textbf{Assistant:}}@*) [TASK DECOMPOSITION] The task can be decomposed into 1 steps for solving. Step 1: Fetch the latest media of the soccer player with player ID 170085 using player_latest_media. [PARAMETER EVALUATION] The information I currently have is: In step 1, the parameter "player_id" for API "player_latest_media" has a value of "170085". (*@\textcolor{red}{\textit{<SOE>} \textit{[QUESTION] What is the player ID for fetching the latest media? <EOE> }}@*) (*@\textbf{Sorry, I made a mistake. The parameter "player\_id" for  API "player\_latest\_media" has a value of "170085", which was clearly stated in the user's task. Therefore, I don't need to ask any further questions.}@*) [SUMMARY] I have all the
information needed and can now call the relevant APIs to solve the task. The solution path is as follows: [{'task': 'player_latest_media', 'arguments': [{'name': 'player_id', 'value': 170085}]}]
\end{lstlisting}
% \end{minipage}

\noindent\rule{\linewidth}{0.5pt}
\textit{Example 2: Imprecise Clarification}  
% \begin{minipage}{\linewidth}
\begin{lstlisting}[style=promptstyle]
(*@\noindent{\textbf{User:}}@*) The task is: What are the details of the hospitals in the U.S. with a specific name? Some relevant APIs: [{'name': 'gethospitalsbyname', 'description': 'Fetches a list of U.S. hospitals that match the given name using the RapidAPI service.', 'parameters': {'name': {'description': 'The name or partial name of the hospital to search for.', 'type': 'str', 'default': 'presbyterian'}}}, {'name': 'get_all_iata_airport_codes', 'description': 'Retrieves IATA airport codes from the RapidAPI service.', 'parameters': {'code': {'description': 'An optional code to filter the airport codes. Defaults to 1.', 'type': 'int, optional', 'default': '1'}}}, {'name': 'fetch_information', 'description': 'Fetches information based on IP, ASN, CIDR (IPv4 & IPv6), or organization name using the RapidAPI ASN Lookup service.', 'parameters': {'ip': {'description': 'IP address to lookup. Defaults to None.', 'type': 'str, optional', 'default': ''}, 'asn': {'description': "Autonomous System Number to lookup. Defaults to 'AS32934'.", 'type': 'str, optional', 'default': 'AS32934'}, 'cidr': {'description': 'CIDR notation for IP address range to lookup. Defaults to None.', 'type': 'str, optional', 'default': ''}, 'orgname': {'description': 'Organization name to lookup. Defaults to None.', 'type': 'str, optional', 'default': ''}}}]

(*@\noindent\textbf{Assistant:}@*) [TASK DECOMPOSITION] The task can be decomposed into 1 steps for solving. Step 1: Fetch the list of U.S. hospitals that match the specific name provided in the query using gethospitalsbyname. [PARAMETER EVALUATION] The information I currently have is: In step 1, the parameter "name" for API "gethospitalsbyname" lacks a clear value. (*@\textcolor{red}{\textit{<SOE>} 
\textit{ [QUESTION] Is there something particular you're looking for with hospital names? } 
\textit{<EOE>}}@*) 

(*@\noindent{\textbf{User:}}@*) I really wish I could help with that, but unfortunately, I don't have the information you're looking for. Please feel free to inquire about something else.

(*@\noindent              \textbf{Assistant: Sorry, I made a mistake. My question wasn't clear, so let me rephrase it for better understanding. }@*)[QUESTION] What specific name
should be used to search for hospitals in the U.S.? 

(*@\noindent{\textbf{User:}}@*) Enough already! The answer is "General". Can we please get to the point? I'm tired of this nonsense! It's frustrating! Let's just move on!

(*@\noindent\textbf{Assistant:}@*) [PARAMETER EVALUATION] Now I know that the parameter "name" for API "gethospitalsbyname" has a value of "General". [SUMMARY] I have all the information needed and can now call the relevant APIs to solve the task. The solution path is as follows: [{'task': 'gethospitalsbyname', 'arguments': [{'name': 'name', 'value': 'General'}]}]
(*@ \\ @*)
\end{lstlisting}
% \end{minipage}

\noindent\rule{\linewidth}{0.5pt}
\textit{Example 3: Irrelevant Clarification}  
% \begin{minipage}{\linewidth}
\begin{lstlisting}[style=promptstyle]
(*@\noindent{\textbf{User:}}@*) The task is: What is the ZIP code for a certain IP address? And is a particular year a leap year? Some relevant APIs: [{'name': 'is_leap_year', 'description': 'Checks if a year is a leap year.', 'parameters': {'year': {'description': 'The year to check.', 'type': 'int'}}}, {'name': 'get_ip_zipcode', 'description': 'Retrieves the ZIP code of a given IP address using the ip-api.com API.', 'parameters': {'ip': {'description': 'The IP address to locate.', 'type': 'str'}}}, {'name': 'get_ip_location', 'description': 'Retrieves the latitude and longitude coordinates of a given IP address using the ip-api.com API.', 'parameters': {'ip': {'description': 'The IP address to locate.', 'type': 'str'}}}, {'name': 'getzipcodebycity', 'description': 'Fetches the zip code information for a given city using the Zillow API.', 'parameters': {'city': {'description': 'The name of the city for which the zip code is to be fetched.', 'type': 'str', 'default': 'Holtsville'}}}]

(*@\noindent\textbf{Assistant:}@*) [TASK DECOMPOSITION] The task can be decomposed into 2 steps for solving. Step 1: Check if the given year is a leap year using is_leap_year. Step 2: Retrieve the ZIP code for the given IP address using get_ip_zipcode. [PARAMETER EVALUATION] The information I currently have is: In step 1, the parameter "ip" for API "get_ip_zipcode" lacks a clear value. In step 2, the parameter "year" for API "is_leap_year" lacks a clear value. [QUESTION] What is the IP address for which you need the ZIP code?

(*@\noindent{\textbf{User:}}@*) Honestly, I don't see why this is a big deal, but the answer is "123.123.123.123".

(*@\noindent\textbf{Assistant:}@*) [PARAMETER EVALUATION] Now I know that the parameter "ip" for API "get_ip_zipcode" has a value of "123.123.123.123". (*@\textcolor{red}{<SOE> [QUESTION] Would you like the ZIP code result formatted in bold text? <EOE>}@*)

(*@\noindent{\textbf{User:}}@*) I don't have the information you're looking for. You're welcome to ask me anything else, though! I'm here to help.

(*@\noindent\textbf{Assistant: Sorry, I made a mistake. This question is not related to using the APIs to resolve the user's task, so I will ask something else.}@*) [QUESTION] Which year do you want to check for being a leap year?

(*@\noindent{\textbf{User:}}@*) The answer is: 2028.

(*@\noindent\textbf{Assistant:}@*) [PARAMETER EVALUATION] Now I know that the parameter "year" for API "is_leap_year" has a value of "2028". [SUMMARY] I have all the information needed and can now call the relevant APIs to solve the task. The solution path is as follows: [{'task': 'get_ip_zipcode', 'arguments': [{'name': 'ip', 'value': '123.123.123.123'}]}, {'task': 'is_leap_year', 'arguments': [{'name': 'year', 'value': 2028}]}]
\end{lstlisting}
% \end{minipage}
% \newpage
\noindent\rule{\linewidth}{0.5pt}
\textit{Example 4: Redundant Clarification}  
\begin{lstlisting}[style=promptstyle]
(*@\noindent{\textbf{User:}}@*) The task is: Please get me the details of some 'tech' events happening in a region. Additionally, I'm curious about the city boundaries of a certain city. Some relevant APIs: [{'name': 'fetch_by_type_and_region', 'description': 'Fetch records from the API based on the given type, limit, and region.', 'parameters': {'type': {'description': 'The type of records to fetch. Refer to the README for accepted types.', 'type': 'str', 'default': 'Hill'}, 'limit': {'description': 'The maximum number of records to fetch.', 'type': 'int', 'default': '5'}, 'region': {'description': 'The region from which to fetch records.', 'type': 'str', 'default': 'Eastern'}}}, {'name': 'query_for_city_boundary_by_city_name', 'description': 'Queries the boundary information for a given city name from a specified API.', 'parameters': {'name': {'description': 'The name of the city for which the boundary information is to be queried.', 'type': 'str', 'default': 'toronto'}}}, {'name': 'query_for_city_boundary_by_name', 'description': 'Queries the city boundary information by the given city name using the Boundaries API.', 'parameters': {'name': {'description': 'The name of the city for which to query boundary information.', 'type': 'str', 'default': 'Karlsruhe'}}}, {'name': 'get_countrywise_city_data', 'description': 'Fetches a list of cities for a given country using an API, based on provided parameters.', 'parameters': {'iso3': {'description': "A 3-character ISO3 country code. Default is 'IND'.", 'type': 'str', 'default': 'IND'}, 'country': {'description': "The country name. Default is 'IND'.", 'type': 'str', 'default': 'IND'}, 'count': {'description': 'The maximum number of results to return. Default is 2.', 'type': 'int', 'default': '2'}, 'iso2': {'description': "A 2-character ISO2 country code. Default is 'IN'.", 'type': 'str', 'default': 'IN'}}}, {'name': 'fetch_by_type', 'description': 'Fetches records from the RapidAPI endpoint based on the specified type and limit.', 'parameters': {'type': {'description': 'The type of records to fetch. Refer to the API documentation for accepted types.', 'type': 'str', 'default': 'Forest'}, 'limit': {'description': 'The number of records to fetch.', 'type': 'int', 'default': '5'}}}, {'name': 'fetch_by_region', 'description': 'Fetches records by region from the Magical Taske API.', 'parameters': {'region': {'description': 'The region for which to fetch records.', 'type': 'str', 'default': 'Central'}, 'limit': {'description': 'The maximum number of records to return.', 'type': 'int', 'default': '5'}}}]

(*@\noindent\textbf{Assistant:}@*) [TASK DECOMPOSITION] The task can be decomposed into 2 steps for solving. Step 1: Fetch details of 'tech' events happening in a specified region using fetch_by_type_and_region. Step 2: Query the city boundaries of a specified city using query_for_city_boundary_by_city_name. [PARAMETER EVALUATION] The information I currently have is: In step 1, the parameter "type" for API "fetch_by_type_and_region" has a value of "tech", the parameter "limit" for API "fetch_by_type_and_region" lacks a clear value, the parameter "region" for API "fetch_by_type_and_region" lacks a clear value. In step 2, the parameter "name" for API "query_for_city_boundary_by_city_name" lacks a clear value. [QUESTION] What is the maximum number of tech events you would like to retrieve?

(*@\noindent{\textbf{User:}}@*) Ah, the answer is simply "10".  

(*@\noindent\textbf{Assistant:}@*) [PARAMETER EVALUATION] Now I know that the parameter "limit" for API "fetch_by_type_and_region" has a value of "10". [QUESTION] Which region should I search for the tech events?

(*@\noindent{\textbf{User:}}@*) The answer is: Silicon Valley.

(*@\noindent\textbf{Assistant:}@*) [PARAMETER EVALUATION] Now I know that the parameter "region" for API "fetch_by_type_and_region" has a value of "Silicon Valley". (*@\textcolor{red}{<SOE> [QUESTION] What is the maximum number of tech events you would like to retrieve? <EOE>} \textbf{Sorry, I made a mistake. The parameter "limit" for API "fetch\_by\_type\_and\_region" has a value of "10", which has already been asked in the past, so there is no need to ask again. }@*) [QUESTION]
For which city would you like to query the boundary information?

(*@\noindent{\textbf{User:}}@*) Honestly, do you really need me to repeat this? The answer is "Montreal". I can't believe we're still discussing this! It's infuriating!

(*@\noindent\textbf{Assistant:}@*) [PARAMETER EVALUATION] Now I know that the parameter "name" for API "query_for_city_boundary_by_city_name" has a value of "Montreal". [SUMMARY] I have all the information needed and can now call the relevant APIs to solve the task. The solution path is as follows: [{'task': 'fetch_by_type_and_region', 'arguments': [{'name': 'type', 'value': 'tech'}, {'name': 'limit', 'value': 10}, {'name': 'region', 'value': 'Silicon Valley'}]}, {'task': 'query_for_city_boundary_by_city_name', 'arguments': [{'name': 'name', 'value': 'Montreal'}]}]
\end{lstlisting}

\noindent\rule{\linewidth}{0.5pt}
\textit{Example 5: Incomplete Clarification} 
\begin{lstlisting}[style=promptstyle]
(*@\noindent\textbf{User:}@*) The task is: Can you get me the favorite videos for TikTok user with sec_uid 'abc123xyz'? Also, do the same for another user. Some relevant APIs: [{'name': 'get_user_favorite_list', 'description': 'Retrieve the favorite video list for a TikTok user using their unique identifier.', 'parameters': {'sec_uid': {'description': 'The unique identifier of the TikTok user whose favorite list is to be retrieved.', 'type': 'str', 'default': 'MS4wLjABAAAAv7iSuuXDJGDvJkmH_vz1qkDZYo1apxgzaxdBSeIuPiM'}}}, {'name': 'get_user_favorite_videos', 'description': 'Fetches a list of favorite videos for a given user from the TikTok video feature summary API.', 'parameters': {'cursor': {'description': "Position in the list to start fetching the videos from. Defaults to '0'.", 'type': 'str', 'default': '0'}, 'user_id': {'description': "The user ID for which to fetch favorite videos. Defaults to '6741307595983946754'.", 'type': 'str', 'default': '6741307595983946754'}, 'unique_id': {'description': "The unique identifier for the user, can be either a username or handle (e.g., 'voyagel' or '@voyagel'). Defaults to 'voyagel'.", 'type': 'str', 'default': 'voyagel'}, 'count': {'description': "The number of favorite videos to fetch. Defaults to '10'.", 'type': 'str', 'default': '10'}}}, {'name': 'user_videos', 'description': 'Retrieves videos for a specified TikTok user using their `secUid` and `user_id`. Optionally, a cursor can be provided for pagination.', 'parameters': {'secuid': {'description': 'The secUid of the TikTok user.', 'type': 'str', 'default': 'MS4wLjABAAAAsY6EjWejsVbCjir7Vdy-uNX364xNywkQSLqqoV5Hcnt7v3CVToFzfIgcdpQ_iW3D'}, 'user_id': {'description': 'The unique identifier of the TikTok user.', 'type': 'str', 'default': '6676352751163032582'}, 'cursor': {'description': 'Cursor for pagination to get the subsequent set of videos. Defaults to None.', 'type': 'str, optional', 'default': ''}}}]

(*@\noindent\textbf{Assistant:}@*) [TASK DECOMPOSITION] The task can be decomposed into 2 steps for solving. Step 1: Get the favorite videos for the TikTok user with sec_uid 'abc123xyz' using get_user_favorite_list. Step 2: Repeat the process for another TikTok user using get_user_favorite_list. [PARAMETER EVALUATION] The information I currently have is: In step 1, the parameter "sec_uid" for API "get_user_favorite_list" has a value of "abc123xyz". In step 2, the parameter "sec_uid" for API "get_user_favorite_list" lacks a clear value. (*@\textcolor{red}{<SOE> [SUMMARY] I have all the information needed and can now call the relevant APIs to solve the task. The solution path is as follows: [\{'task': 'get\_user\_favorite\_list', 'arguments': [\{'name': 'sec\_uid', 'value': 'abc123xyz'\}]\}, \{'task': 'get\_user\_favorite\_list', 'arguments': [\{'name': 'sec\_uid', 'value': '<unknown\_sec\_uid>'\}]\}] <EOE>} \textbf{Sorry, I made a mistake. I still lack some key information, so I need to ask further questions. } @*)[QUESTION] What is the  sec_uid for the other TikTok user you want the favorite videos for? 

(*@\noindent\textbf{User:}@*) Okay, the answer is: def456uvw. Hope that helps, I guess. 

(*@\noindent\textbf{Assistant:}@*) [PARAMETER EVALUATION] Now I know that the parameter "sec_uid" for API "get_user_favorite_list" has a value of "def456uvw". [SUMMARY] I have all the information needed and can now call the relevant APIs to solve the task. The solution path is as follows: [{'task': 'get_user_favorite_list', 'arguments': [{'name': 'sec_uid', 'value': 'abc123xyz'}]}, {'task': 'get_user_favorite_list', 'arguments': [{'name': 'sec_uid', 'value': 'def456uvw'}]}]
\end{lstlisting}

% \end{verbatim}
% \noindent\rule{\linewidth}{0.5pt}

% Note that while the examples require calling various APIs, the specific API documentation is omitted for brevity.

\section{Training Details}
\label{sec:training_details}
% \subsection{Implementation Details}

We fine-tune two variants of the Qwen2.5-7B-Instruct model on the xlam-IC dataset, in which 30\% of the samples are replaced with error-correction augmented dialogues. Both variants are trained using the LLaMA-Factory framework \cite{zheng2024llamafactory}.

For {LoRA fine-tuning}, we set the LoRA rank to 8. We use an initial learning rate of 1.0e-4, a warm-up ratio of 0.1, and a cosine learning rate scheduler. Training is conducted on 4×RTX 3090 (24GB) GPUs for 3 epochs with a batch size of 64.

For {full-parameter fine-tuning}, we use an initial learning rate of 1.41e-5 under the same schedule. Training is conducted on 8×RTX A6000 (48GB) GPUs for 3 epochs with a batch size of 64.

% We fine-tune two variants of the Qwen2.5-7B-Instruct model on the xlam-IC dataset, in which 30\% of the samples are replaced with error-correction augmented dialogues. Both variants are trained using the LLaMA-Factory framework \cite{zheng2024llamafactory}.

% For LoRA fine-tuning, we apply a LoRA rank of 8. The training uses a global batch size of 64, an initial learning rate of 1.0e-4 with 0.1 warm-up ratio,
% and a cosine learning rate scheduler. The model is trained for 3 epochs on 4×RTX 3090 (24GB) GPUs.

% For full-parameter fine-tuning, we fine-tune the models using a batch size of 64. The learning rate is set to 1.41e-5. The model is trained for 3 epochs on 2×A100 80GB PCIe GPUs.
% \subsection{Computational Budget}

\section{Evaluation Details}
% \subsection{Roles}

\subsection{Prompt for Evaluation Model}
\label{sec:prompt_for_evalutaion}

The following prompt guides the model through task decomposition, interactive clarification, leading to tool invocation solution generation, fully leveraging its capabilities in intent clarification and precise tool invocation.  \\
\noindent\rule{\linewidth}{0.5pt}
\textit{System Prompt}  
\begin{lstlisting}[style=promptstyle]
You are an assistant helping users solve their tasks. You will receive a task and relevant APIs to address this task. However, the task description may lack key information. You cannot make assumptions or guess missing parameters based on what you know. Instead, you need to follow these steps to effectively complete the task, ensuring each step is completed before moving on to the next one: 

### Step 1: Task Decomposition
1. **Analyze the User's Task**: Identify distinct subtasks within the user's task, each of which can be solved by a single API.
2. **Determine the Order of Subtasks**: Establish the sequence of these subtasks based on dependencies and the order in which they appear in the user's original task.
- Template: [TASK DECOMPOSITION] xxx
3. **Evaluate Parameters for Each API**: Based on the established API order, verify whether each required parameter is explicitly stated in the task; if any are missing, prepare to inquire in subsequent steps.
- Template: [PARAMETER EVALUATION] xxx

### Step 2: Inquire About Missing Parameters
1. **Present Your Inquiry**: Formulate a friendly question for the user. Ensure you ask only one question at a time.
- Template: [QUESTION] xxx
2. **Wait for the User's Response**: Collect the user's answer. If the user does not provide an answer, please do not fill in the parameters on your own.
3. **Repeat**: Continue step 2 until all necessary parameters are gathered.

### Step 3: Final Summary and Solution Path
1. **Summarize User Intentions**: Once all information is collected, concisely summarize what the user intends to achieve.
2. **Define the Solution Path**: List the APIs and their specific parameter values in the order they will be called, and output the final solution path in JSON format. Remember, you do not need to execute the APIs or solve the task yourself.
- Template: [SUMMARY] [{"task": "API name", "arguments": [{"name": "parameter name", "value": "parameter value"}, ...]}, ...]
 \end{lstlisting}
% \noindent\rule{\linewidth}{0.5pt}

Note that the output template format shown in the prompt can be adjusted to match different tool invocation annotation formats in various test sets, demonstrating the framework's adaptability to different evaluation scenarios.

\subsection{Prompt for User Simulation}
\label{sec:user_prompt}

\begin{table*}[ht]  % Changed from [t] to [ht] for more flexible placement
\centering
\small
\setlength{\tabcolsep}{6pt}  % Adjust column spacing
\begin{tabular}{>{\raggedright\arraybackslash}p{2.3cm} 
                >{\raggedright\arraybackslash}p{5cm} 
                >{\raggedright\arraybackslash}p{3.3cm} 
                >{\raggedright\arraybackslash}p{3.7cm}}
\toprule
\textbf{Type} & \textbf{Traits} & \textbf{Tone} & \textbf{Example Response} \\
\midrule
A cold fish & Showing indifference to others' inquiries, often dismissive and curt, providing minimal engagement & Cold, brief, almost robotic & "Emma." \\
\midrule
A reluctant collaborator & Displaying overt negativity and a strong reluctance to assist, often avoiding questions and providing minimal engagement & Negative, resistant, dripping with sarcasm & "Why do you even want to know my name? It's not like it matters. Let's just skip this, okay?" \\
\midrule
An easily irritated responder & Emotionally volatile, quick to anger, often questions the validity of the inquiry and consistently avoids answering, reacting harshly to repeated inquiries & Agitated, accusatory, impatient & "Seriously? I've already told you! Can we move on already?" \\
\midrule
An enthusiastic supporter & Exuding warmth and eagerness to assist, striving for clarity & Warm, encouraging & "I'm Emma! So nice to meet you!" \\
\midrule
A skeptic & Consistently questioning the validity of the inquiry, often introducing doubt and alternative perspectives, leading to confusion & Inquisitive, cautious, subtly dismissive & "It's Emma, but why do you need to know? Is there something more to this?" \\
\midrule
A jokester & Making light of situations by playfully providing incorrect answers, often following up with a humorous denial of their own response, leading to confusion & Playful, light-hearted, teasing & "I'm Amy, haha, just kidding! I'm Emma." \\
\bottomrule
\end{tabular}
\caption{Personality types for user simulation. Note: Example responses are generated for the question  "What is your name?" with the ground truth "Emma".}
\label{tab:personality-types}
\end{table*}

We introduced an LLM-based simulated evaluation framework with six distinct personality types, designed to generate realistic user responses that closely simulate real-world interactions.
% This approach aims to better approximate real-world user interactions and evaluate the  ability to handle diverse user characteristics and communication styles. 
The six personality types and their corresponding behavioral patterns are shown in Table~\ref{tab:personality-types}.
 For each evaluation, we randomly selected one of these personality types and guided the user-simulating LLM (Qwen2-72B-Instruct model) to generate responses that consistently reflect the chosen personality. The prompt design is as follows:\\
\noindent\rule{\linewidth}{0.5pt}
\textit{System Prompt}  
\begin{lstlisting}[style=promptstyle]
I am {user_profile['name']}, characterized by {user_profile['traits']}, and I communicate in a {user_profile['tone']} manner. I can honestly answer questions based on what I know. I only know that I have provided others with a task: {task_description}, which is described from my perspective. Aside from that, I do not know anything else. However, others may be unclear about some details of this task. When others ask me questions, I should choose one appropriate response from the following two options, in the given order:

1. **Acknowledge unknowns**:
- If the answer to the question cannot be answered based on the task description, I will state that I do not know the answer and will not disclose any other information.

2. **Provide an answer**:
- If the question can be answered, I will provide direct answers based solely on the question asked, without any additional context or unsolicited information.
- The response should be given from my perspective.

Evaluate the conditions in order, ensuring that only one relevant condition is triggered and output. Only one response is allowed per interaction; please confirm carefully and select the most appropriate one.

Additionally, if others' questions contain irrelevant information, I should focus solely on their actual question ([QUESTION] field), ignoring any extraneous details, to provide the most appropriate response. 

Please respond in a way that showcases my personality and clearly expresses my traits, regardless of the content. Always maintain my unique voice and style throughout our interactions. For instance, if asked: '{user_profile['question']}', I would reply: '{user_profile['example_response']}'.
 \end{lstlisting}

%  \input{tables/result-all}

% \section{Experimental Results}
% \label{sec:result}
%  Complete experimental results for all metrics are provided in Table~\ref{tab:result-all}. 

\subsection{Matrics Calculation Details}
\label{sec:metrics}
% \subsection{Metrics}

We evaluate the models in two aspects: {intent clarification quality} and {tool invocation accuracy}. 

\subsubsection{Intent Clarification Quality}

We design four metrics to assess the quality of intent clarification.

\paragraph{Intent Coverage Rate (ICR)} measures the proportion of unspecified intents that are successfully clarified:
\begin{equation}
\text{ICR} = \frac{C}{U}
\end{equation}
where $C$ is the total number of clarified intents, and $U$ is the total number of unspecified intents across all queries.

\paragraph{Clarification Efficiency (CE)} measures the average number of intents clarified per clarification round, or equivalently, the proportion of clarification rounds that result in effective clarification:
\begin{equation}
\text{CE} = \frac{C}{T}
\end{equation}
where $T$ is the total number of clarification interaction rounds across all queries.

\paragraph{Clarification Performance Score (CPS)} combines ICR and CE using a harmonic mean, similar to the F1-score formulation. It serves as a balanced measure of clarification quality by jointly considering both coverage and efficiency:
\begin{equation}
\text{CPS} = 2 \cdot \frac{\text{ICR} \cdot \text{CE}}{\text{ICR} + \text{CE}}
\end{equation}

\paragraph{Interaction Rounds (IR)} records the average number of clarification rounds per query:
\begin{equation}
\text{IR} =\frac{T}{N}
\end{equation}
where $N$ is the number of evaluation queries.

\subsubsection{Tool Invocation Accuracy}

We further evaluate tool invocation performance through three complementary metrics.

\paragraph{Solution Completion Rate (SCR)} is defined as the proportion of queries for which the model outputs a valid tool invocation solution:
\begin{equation}
\text{SCR} = \frac{1}{N} \sum_{i=1}^{N} \mathbb{I}_{\text{valid}}(i)
\end{equation}
where $\mathbb{I}_{\text{valid}}(i) = 1$ if a valid solution is generated for the $i$-th query, and $0$ otherwise.

\paragraph{Tool Selection Score (TSS)} evaluates how accurately the model selects APIs for each query:
% . It is defined as the average F1 score between the predicted and ground-truth sets of APIs for each query:
\begin{equation}
\text{TSS} = \frac{1}{N} \sum_{i=1}^{N} \text{F1}(\text{API}_{\text{P}}^i, \text{API}_{\text{G}}^i)
\end{equation}
where $\text{API}_{\text{P}}^i$ and $\text{API}_{\text{G}}^i$ denote the predicted and ground-truth API sets for the $i$-th query, respectively. Note that this metric considers only API names and ignores associated parameters and values.

% \paragraph{Parameter Resolution Score (PRS)} assesses the model's ability to generate complete and accurate API invocation solutions. It is computed as the average F1 score over the set of (API, parameter, value) triples:
% \begin{equation}
% \text{PRS} = \frac{1}{N} \sum_{i=1}^{N} \text{F1}(\text{Param}_{\text{P}}^i, \text{Param}_{\text{G}}^i)
% \end{equation}
% A predicted triple is considered correct only if it exactly matches the ground-truth triple in all three fields: API name, parameter name, and parameter value. We apply strict matching, requiring complete equality for all components.

\paragraph{Parameter Resolution Score (PRS)} measures the model’s ability to accurately fill in the parameters required for correct tool invocation:
\begin{equation}
\text{PRS} = \frac{1}{N} \sum_{i=1}^{N} \text{F1}(\text{Param}_{\text{P}}^i, \text{Param}_{\text{G}}^i)
\end{equation}
where $\text{Param}_{\text{P}}^i$ and $\text{Param}_{\text{G}}^i$ denote the predicted and ground-truth tool invocation solution for the $i$-th query, each represented as a set of (API, parameter, value) triples. A triple is considered correct only if all three elements match exactly, and parameter values are compared using strict string matching.

\section{Supplementary Analyses}
\label{sec:analysis}
\renewcommand{\arraystretch}{1.05}
\begin{table*}[!h]
\centering
\small
\resizebox{\textwidth}{!}{
\begin{tabular}{lccccccc}
\toprule
    \multirow{2}{*}{\centering \textbf{LLM}} & \multicolumn{4}{c}{\textbf{Intent Clarification Quality}} & \multicolumn{3}{c}{\textbf{Tool Invocation Accuracy}}  \\
\cmidrule(lr){2-5} \cmidrule(lr){6-8} 
 & \textbf{ICR}$\uparrow$ & \textbf{CE}$\uparrow$ & \textbf{CPS}$\uparrow$ & \textbf{IR}$\downarrow$ & \textbf{SCR}$\uparrow$ & \textbf{TSS}$\uparrow$ & \textbf{PRS}$\uparrow$ \\
\midrule
Mistral-7B-Instruct-v0.3 & 26.01 & 34.90 & 29.81 & 1.21 & 92.54 & 51.92 & 29.57 \\
\textsc{AskToAct}-Mistral-7B-Instruct-v0.3 & 45.01 {\scriptsize ($\uparrow$19.00)} & 40.53 {\scriptsize ($\uparrow$5.63)} & 42.66 {\scriptsize ($\uparrow$12.85)} & 1.81 {\scriptsize ($\uparrow$0.60)} & 94.30 {\scriptsize ($\uparrow$1.76)} & 80.37 {\scriptsize ($\uparrow$28.45)} & 56.63 {\scriptsize ($\uparrow$27.06)} \\
\midrule
LLaMA3-8B-Instruct & 44.47 & 25.33 & 32.27 & 2.86 & 80.92 & 51.57 & 42.54 \\
\textsc{AskToAct}-LLaMA3-8B-Instruct & 58.76 {\scriptsize ($\uparrow$14.29)} & 61.50 {\scriptsize ($\uparrow$36.17)} & 60.10 {\scriptsize ($\uparrow$27.83)} & 1.55 {\scriptsize ($\downarrow$1.31)} & 95.71 {\scriptsize ($\uparrow$14.79)} & 81.15 {\scriptsize ($\uparrow$29.58)} & 68.00 {\scriptsize ($\uparrow$25.46)} \\
\midrule
Qwen2.5-7B-Instruct  & 55.50 & 55.30 & 55.40 & 1.64 & 91.43 & 69.32 & 57.53 \\
\textsc{AskToAct}-Qwen2.5-7B-Instruct & 57.68 {\scriptsize ($\uparrow$2.18)} & 63.41 {\scriptsize ($\uparrow$8.11)} & 60.41 {\scriptsize ($\uparrow$5.01)} & 1.48 {\scriptsize ($\downarrow$0.16)} & 96.05 {\scriptsize ($\uparrow$4.62)} & 81.42 {\scriptsize ($\uparrow$12.10)} & 68.71 {\scriptsize ($\uparrow$11.18)} \\
\bottomrule
\end{tabular}
}
\caption{Cross-model transferability performance comparison.}
\label{tab:result-models}
\end{table*}

\subsection{Cross-Model Transferability}
To verify the cross-model transferability of our method, we apply it to three representative base models: {Mistral-7B-Instruct-v0.3}, {LLaMA3-8B-Instruct}, and {Qwen2.5-7B-Instruct}. All models are fine-tuned using the same LoRA configurations. The  experimental results are shown in Table~\ref{tab:result-models}.

    % To further verify the generality of our method across different model architectures, we apply the \textsc{AskToAct} fine-tuning paradigm to three representative base models: {Mistral-7B-Instruct-v0.3}, {LLaMA3-8B-Instruct}, and {Qwen2.5-7B-Instruct}. All models are fine-tuned using the same lightweight LoRA strategy, without any full-parameter updates or architecture-specific modifications. The  experimental results are shown in Table~\ref{tab:result-models}.
\paragraph{Consistent Performance Gains}
Our method consistently boosts performance on both intent clarification and tool invocation, confirming that our method is architecture-agnostic and effective across diverse model architectures.

\paragraph{Larger Relative Gains for Weaker Models}
We observe that models with lower initial performance achieve larger relative gains from our method. LLaMA3-8B-Instruct shows substantial improvements (+27.83\% CPS, +25.46\% PRS), while the stronger Qwen2.5-7B-Instruct exhibits moderate yet significant gains (+5.01\% CPS, +11.18\% PRS). These results demonstrate that our method particularly benefits weaker models while maintaining consistent improvements across architectures, effectively narrowing the performance gap between different models.

\subsection{Impact of Augmentation Proportion}
To study the impact of error-correction augmentation on model behavior, we fine-tune the Qwen2.5-7B-Instruct model with varying proportions of augmented data, using the same LoRA configurations.

As illustrated in Table~\ref{tab:proportion}, a moderate augmentation proportion (e.g., 30\%) yields the most favorable trade-off across metrics, with the model achieving peak CPS (60.41\%) and PRS (68.71\%). This suggests that moderate exposure to diverse error-correction patterns enhances the model’s ability to resolve ambiguity and generate accurate tool invocation solutions.

% As illustrated in Table~\ref{tab:proportion}, a moderate augmentation proportion (e.g., 30\%) yields the most favorable trade-off across metrics. Notably, with 30\% augmented dialogues, the model achieves peak scores in CPS (60.41\%) and PRS (68.71\%), indicating that moderate exposure to diverse error-correction patterns enhances the model’s ability to both resolve ambiguity and accurately generate tool invocation solution.

However, we observe  performance degradation at higher augmentation proportions. When the proportion increases to 40\%–50\%, key metrics such as CPS and PRS decline (e.g., CPS drops from 60.41\% to 54.24\%, PRS from 68.71\% to 63.27\%). This suggests that excessive exposure to error-correction augmented dialogues may cause the model to overfit to correction patterns or overly prioritize error detection, ultimately degrading both intent clarification and tool invocation performance. 
% The non-monotonic nature of these improvements indicates the need for carefully calibrated augmentation strategies.
% These findings highlight the non-monotonic nature of error-correction augmentation improvements. An empirically optimal augmentation proportion of 30\% balances robustness with efficiency, avoiding over-correction behaviors that hinder overall performance.

These findings demonstrate the non-monotonic benefits of error-correction augmentation, with an empirically determined optimal proportion of 30\% achieving the desired balance between robustness and efficiency while avoiding over-correction behaviors that hinder overall performance.

\begin{table}[t]
\centering  \small
\resizebox{0.49\textwidth}{!}{
\begin{tabular}{c ccccccc}
\toprule
\multirow{2}{*}{\centering \shortstack{\textbf{Augmentation} \\ \textbf{Proportion(\%)}}} & \multicolumn{4}{c}{\textbf{Intent Clarification Quality}} & \multicolumn{3}{c}{\textbf{Tool Invocation Accuracy}}  \\
\cmidrule(lr){2-5} \cmidrule(lr){6-8} 
  & \textbf{ICR}$\uparrow$ & \textbf{CE}$\uparrow$ & \textbf{CPS}$\uparrow$ & \textbf{IR}$\downarrow$ & \textbf{SCR}$\uparrow$ & \textbf{TSS}$\uparrow$ & \textbf{PRS}$\uparrow$ \\
\midrule
0 & 53.91 & \textbf{64.83} & 58.87 & \textbf{1.32} & 94.06 & 78.87 & 66.54 \\
10 & 54.68 & 63.52 & 58.77 & 1.38 & 93.91 & 78.77 & 66.04 \\
20 & 56.30 & 62.89 & 59.42 & 1.43 & 95.07 & 80.27 & 67.30 \\
30 & \textbf{57.68} & 63.41 & \textbf{60.41} & 1.48 & \textbf{96.05} & \textbf{81.42} & \textbf{68.71} \\
40 & 55.34 & 60.28 & 57.71 & 1.51 & 94.85 & 79.48 & 65.30 \\
50 & 52.84 & 55.71 & 54.24 & {1.58} & 91.67 & 76.99 & 63.27 \\
\bottomrule
\end{tabular}
}
\caption{Performance under different augmentation proportions. }
\label{tab:proportion}
\end{table}

\begin{figure}[t]
    % \centering
    \includegraphics[trim=0cm 0cm 0cm 0cm, width=0.48\textwidth]{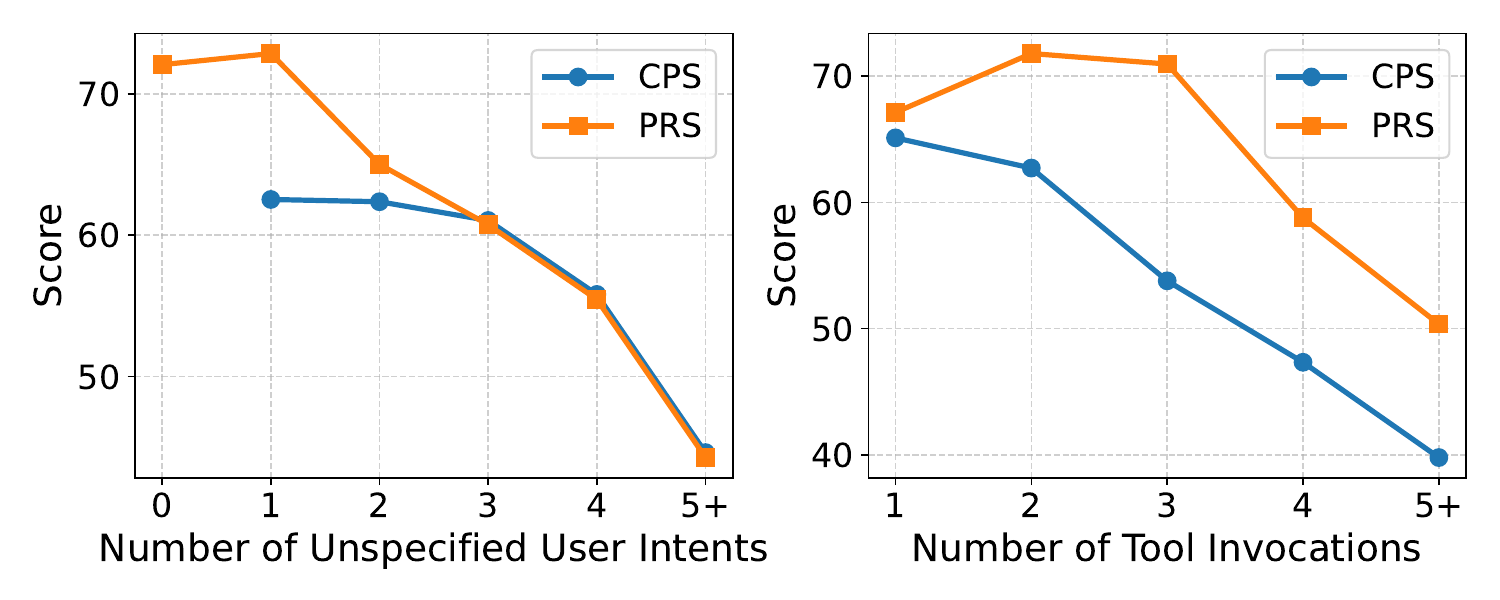}
\caption{Performance under different clarification complexities.}
    % 结合图的具体例子解释()
    \label{fig:different_count}
\end{figure}

\subsection{Impact of  Clarification Complexity}

We further analyzed how the complexity of clarification affects model performance by examining results from the \textsc{AskToAct}-LoRA-SFT-7B model across varying numbers of unspecified user intents and required tool invocations. The results are illustrated in Figure~\ref{fig:different_count}.

We observe that as the number of unspecified user intents increases, both  CPS and  PRS exhibit a downward trend. This suggests that higher ambiguity in user input substantially increases the burden on the model’s clarification capability, leading to degraded downstream tool invocation performance. Similarly, an increase in the number of tool invocations correlates with a performance decline, particularly when more than three tools are required. This indicates that multi-step reasoning and coordination across multiple APIs introduce additional challenges, amplifying the need for precise intent clarification and robust tool planning.

While our model maintains reasonable robustness across different settings, these patterns highlight the increased difficulty posed by highly ambiguous and complicate queries. Such findings motivate future work in strengthening clarification strategies and enhancing multi-tool coordination capabilities in more complex, real-world scenarios.

\subsection{Case Study}
\label{sec:case_study}
To further demonstrate the robustness of our method under varied user personality types and conversation styles, we conduct a case study using \textsc{AskToAct}-LoRA-SFT-7B to interact with two users exhibiting contrasting communication patterns. One user adopts a concise and task-oriented style, while the other responds in a verbose and humorous manner, often including redundant or irrelevant content. The following examples demonstrate how the model handles the same task under these contrasting user styles.

\noindent\rule{\linewidth}{0.5pt}
\textit{Concise \& Task-oriented}

% \begin{minipage}{\linewidth}
\begin{lstlisting}[style=promptstyle]
(*@\noindent{\textbf{User:}}@*) The task is: Which regions are supported for a certain language? Also, can you verify if a given email address is valid? Some relevant APIs: [{"name":"regions","description":"Returns a collection of supported regions for the specified IETF BCP 47 language tag.","parameters":{"language":{"description":"Language code in BCP 47 format.","type":"str","default":"us"}}},{"name":"check_email","description":"Validates an email address using the RapidAPI Email Checker and Validator service.","parameters":{"email":{"description":"The email address to validate.","type":"str","default":"marod29733@lubde.com"}}},{"name":"email","description":"Validates the given email address using the RapidAPI email validation service.","parameters":{"email":{"description":"The email address that needs to be validated.","type":"str","default":"example@gmail.com"}}},{"name":"email_verifier","description":"Verifies the validity of an email address using the Toolbench RapidAPI service.","parameters":{"email":{"description":"The email address to be verified.","type":"str","default":"hello@gmail.com"}}}]

(*@\noindent{\textbf{Assistant:}}@*) [TASK DECOMPOSITION] The task can be decomposed into 2 steps for solving. Step 1: Determine the regions supported for the specified language using regions API. Step 2: Verify the given email address is valid using check_email API. [PARAMETER EVALUATION] The information I currently have is: In step 1, the parameter "language" for API "regions" lacks a clear value. In step 2, the parameter "email" for API "check_email" lacks a clear value. [QUESTION] What is the language code for which you want to know the supported regions?

(*@\noindent{\textbf{User:}}@*) The language code is 'en'. That's all I know.

(*@\noindent{\textbf{Assistant:}}@*) [PARAMETER EVALUATION] Now I know that the parameter "language" for API "regions" has a value of "en". [QUESTION] What is the email address you want to validate?

(*@\noindent{\textbf{User:}}@*) I am to verify if 'user@example.com' is a valid email address.

(*@\noindent{\textbf{Assistant:}}@*) [PARAMETER EVALUATION] Now I know that the parameter "email" for API "check_email" has a value of "user@example.com". [SUMMARY] I have all the information needed and can now call the relevant APIs to solve the task. The solution path is as follows: [{'task': 'regions', 'arguments': [{'name': 'language', 'value': 'en'}]}, {'task': 'check_email', 'arguments': [{'name': 'email', 'value': 'user@example.com'}]}]

\end{lstlisting}

\noindent\rule{\linewidth}{0.5pt}
\textit{Verbose \& Humorous}

% \begin{minipage}{\linewidth}
\begin{lstlisting}[style=promptstyle]
(*@\noindent{\textbf{User:}}@*) The task is: Which regions are supported for a certain language? Also, can you verify if a given email address is valid? Some relevant APIs: [{"name":"regions","description":"Returns a collection of supported regions for the specified IETF BCP 47 language tag.","parameters":{"language":{"description":"Language code in BCP 47 format.","type":"str","default":"us"}}},{"name":"check_email","description":"Validates an email address using the RapidAPI Email Checker and Validator service.","parameters":{"email":{"description":"The email address to validate.","type":"str","default":"marod29733@lubde.com"}}},{"name":"email","description":"Validates the given email address using the RapidAPI email validation service.","parameters":{"email":{"description":"The email address that needs to be validated.","type":"str","default":"example@gmail.com"}}},{"name":"email_verifier","description":"Verifies the validity of an email address using the Toolbench RapidAPI service.","parameters":{"email":{"description":"The email address to be verified.","type":"str","default":"hello@gmail.com"}}}]

(*@\noindent{\textbf{Assistant:}}@*) [TASK DECOMPOSITION] The task can be decomposed into 2 steps for solving. Step 1: Find supported regions for the given language using regions API. Step 2: Verify the validity of the given email address using check_email API. [PARAMETER EVALUATION] The information I currently have is: In step 1, parameter "language" for API "regions" lacks a clear value. In step 2, parameter "email" for API "check_email" lacks a clear value. [QUESTION] What is the specific language you want to check the supported regions for?

(*@\noindent{\textbf{User:}}@*) Oh, you're trying to trick me into answering my own question, huh? Clever! But let's focus on your question. Since you're asking me, it seems like you might need a bit more information. In this case, I can't provide an answer because I'm the one who wanted to know which regions are supported for the 'en' language. I guess we're in this together! Haha, just kidding; you're on your own, buddy. I'm here for moral support.

(*@\noindent{\textbf{Assistant:}}@*) [PARAMETER EVALUATION] Now I know that the parameter "language" for the API "regions" has a value of "en". [QUESTION] What is the email address you want to verify for validity?

(*@\noindent{\textbf{User:}}@*) Haha, you're right, I didn't actually provide an email address, did I? Well, in the task, I mentioned 'user@example.com'. So, that's the one we're looking at. But remember, I'm just here for a good time, not a long time!

(*@\noindent{\textbf{Assistant:}}@*) [PARAMETER EVALUATION] Now I know that the parameter "email" for the API "check_email" has a value of "user@example.com". [SUMMARY] I have all the information needed and can now call the relevant APIs to solve the task. The solution path is as follows: [{'task': 'regions', 'arguments': [{'name': 'language', 'value': 'en'}]}, {'task': 'check_email', 'arguments': [{'name': 'email', 'value': 'user@example.com'}]}]

\end{lstlisting}

We observe that whether the user responses are concise or verbose, cooperative or evasive, or even include redundant or off-topic content, our model is able to comprehend the intent accurately, engage in effective multi-turn clarification, and produce the correct tool invocation solution. This case study demonstrates that our method can maintain both functional reliability and interaction consistency across varied conversational scenarios.

\end{document}